\documentclass{article}

\usepackage{arxiv}

\usepackage[utf8]{inputenc} % allow utf-8 input
\usepackage[T1]{fontenc}    % use 8-bit T1 fonts
\usepackage{hyperref}       % hyperlinks
\usepackage{url}            % simple URL typesetting
\usepackage{booktabs}       % professional-quality tables
\usepackage{amsfonts}       % blackboard math symbols
\usepackage{nicefrac}       % compact symbols for 1/2, etc.
\usepackage{microtype}      % microtypography
\usepackage{lipsum}		% Can be removed after putting your text content
\usepackage{graphicx}
\usepackage{wrapfig, blindtext}
\usepackage{multirow}
\usepackage{amssymb}
\usepackage{subfigure}
\usepackage{graphicx}
\usepackage{amsmath,amssymb,amsfonts}
\usepackage{mathrsfs}
\usepackage{algorithm2e}
\newcommand{\R}{\mathbb{R}}

\newcommand{\bm}{{m}}

 \title{Tensor Decomposition for Multi-agent Predictive State Representation}

\author{ \href{https://orcid.org/0000-0001-5805-072X}{\includegraphics[scale=0.06]{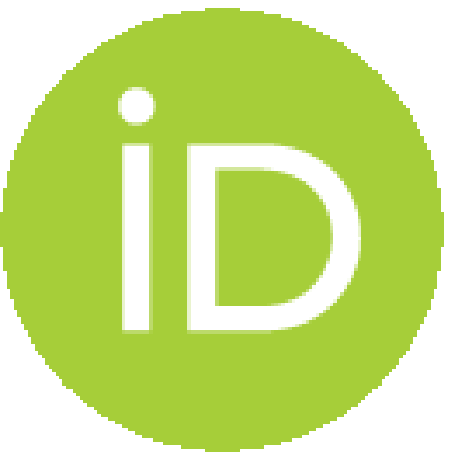}\hspace{1mm}Bilian Chen}\\
	Department of Automation\\ Xiamen University\\ Xiamen 361005\\ China\\
	\texttt{blchen@xmu.edu.cn} \\
	\And
	\href{https://orcid.org/0000-0003-1515-6449}{\includegraphics[scale=0.06]{orcid.eps}\hspace{1mm}Biyang Ma} \\
	School of Computing\\ Teesside University\\ TS1 3BX\\ UK\\
	\texttt{B.Ma@tees.ac.uk} \\
		\And
	\href{https://orcid.org/0000-0002-5246-403X}{\includegraphics[scale=0.06]{orcid.eps}\hspace{1mm}Yifeng Zeng~\thanks{Corresponding author.} } \\
School of Computing\\ Teesside University\\ TS1 3BX\\ UK\\
\texttt{y.zeng@tees.ac.uk} \\
		\And
\href{https://orcid.org/0000-0001-6388-1413}{\includegraphics[scale=0.06]{orcid.eps}\hspace{1mm}Langcai Cao}\\
Department of Automation\\ Xiamen University\\ Xiamen 361005\\ China\\
\texttt{langcai@xmu.edu.cn} \\
\And
\href{https://orcid.org/0000-0000-0000-0000}{\includegraphics[scale=0.06]{orcid.eps}\hspace{1mm}Jing Tang} \\
School of Computing\\ Teesside University\\ TS1 3BX\\ UK\\
\texttt{j.tang@tees.ac.uk} \\
}

\hypersetup{
pdftitle={Tensor Decomposition for Multi-agent Predictive State Representation},
pdfsubject={cs.MA},
pdfauthor={Blian~Chen,Biyang~Ma, Yifeng~Zeng, Langcai~Cao, Jing~Tang},
pdfkeywords={ Predictive state representations, Tensor optimization, Learning approaches},
}

\begin{document}
\maketitle

\begin{abstract}
Predictive state representation~(PSR) uses a vector of action-observation sequence to represent the system dynamics and subsequently predicts the probability of future events.  It is a concise knowledge representation that is well studied in a  single-agent planning problem domain. To the best of our knowledge, there is no existing work on using PSR to solve multi-agent planning problems. Learning a multi-agent PSR model is quite difficult especially with the increasing number of agents, not to mention the complexity of a problem domain. In this paper, we resort to tensor techniques to tackle the challenging task of multi-agent PSR model development problems.
By first focusing on a two-agent setting, we construct the system dynamics matrix as a high order tensor for a PSR model, learn the prediction parameters and deduce state vectors directly through two different tensor decomposition methods respectively, and derive the transition parameters via linear regression. Subsequently, we generalize the PSR learning approaches in a multi-agent setting. Experimental results show that our methods can effectively solve multi-agent PSR modelling problems in multiple problem domains.
\end{abstract}

% keywords can be removed

  \section{Introduction}

 Predictive State Representation~(PSR) is a dynamic system modelling method and uses a vector of action-observation sequence to represent system states, which is subsequently used to solve a sequence prediction problem \cite{Littman2001Predictive}. The system dynamics matrix theory provides a matrix-based modelling technique for learning PSR \cite{Singh2004Predictive}.
 Currently the PSR discovery and learning algorithms have been well studied except that the algorithmic reliability and efficiency needs to be improved, e.g., the search based techniques \cite{James2004Learning,Wolfe2005Learning,Huang2018Basis}, the spectral learning approach \cite{Boots2011An,Kulesza2015Spectral}, the compressed sensing approach \cite{Hamilton2014Efficient,Hamilton2014Modelling} and the sub-state space method \cite{Liu2015Predictive}. However, the PSR research is solely conducted in a single-agent decision making setting.
 % and these tasks are only modelled on a single agent.
 %Many efforts have focused on single agent situation and have shown expected performance, e.g., the search based techniques  \cite{James2004Learning,Wolfe2005Learning,Liu2015Predictive,Huang2018Basis}, spectral learning approach\cite{Boots2011An,Kulesza2015Spectral}, compressed sensing approach \cite{Hamilton2014Efficient,Hamilton2014Modelling},and sub-state space method \cite{Liu2015Predictive}.
 
 %,optimization technique\cite{Zeng2017Group} and planning problem\cite{James2004Planning,James2004Planningin,Boots2010Closing,Boularias2010Predictive,Hamilton2014Efficient}
 
 Learning a multi-agent PSR model is rather difficult since available data contains interactive behaviour of multiple agents, e.g. their observations and actions, and the interaction data is often to be considered as a high dimensional space particularly with the increasing number of agents.
 %especially  when the number of agent is increasing, not to mention the complexity of problem domain is aggravated.
 Moreover, the multi-agent  system dynamics matrix is often filled with noise and when the available data is not sufficient in a complicated problem domain, it would be hard to learn a good PSR model  in a large multi-agent problem domain.
 Meanwhile, the computational cost will be dramatically increased since a large number of tests need to be conducted in order to build high dimensional matrices for learning a multi-agent PSR model.
 %
 % too many tests will cause too many manipulations in constructing high dimensional matrixes, thus the computational cost may be too high to afford.
 %In summary, the current PSRs modelling research has not covered the multi-agent modelling problem, and most of the research work still focuses on the improvement of model learning efficiency and scalability.
 %Therefore, this paper mainly promotes the PSRs model of multi-agent based on the PSRs model of single agent, and solves the challenging multi-agent modelling problem, especially the PSRs model from a large number of multi-agent interaction data.

 In this paper, we focus on a PSR model with more than one agent and investigate a high dimensional system dynamics matrix, namely tensor, for learning the multi-agent PSR model.
 %, and learn the PSR models through tensor decomposition.
 The key underlying idea is to take advantage of a highly connected structure of a tensor and the property of tensor decomposition that
 can extract the latent low-rank components~(even though the data is noisy).
 The difficulty lies in the embedding of tensor into the multi-agent PSR model and the learning of model prediction parameters, state vectors and prediction equation.
 We present two commonly used tensor decomposition techniques~(CP decomposition and its generalized form Tucker decomposition) to solve the PSR discovery and learning problems.
 Thus,  the model prediction parameters and the compressed vector for representing states can be obtained from the decomposition results.
 Inspired by the transformed PSR model \cite{Rosencrantz2004Learning}, we obtain the model transition parameters  via linear regression after constructing auxiliary matrices. We conduct experiments on several problem domains including one extremely large domain,
 and the results demonstrate the expected performance.

 The rest of this paper is organized as follows.
 In Section \ref{sec:preparations}, we extend  a single-agent PSR model, which leads to a multi-agent PSR model, to represent a multi-agent planning problem.
 Section \ref{sec:SDT} introduces system dynamics tensor for learning a multi-agent PSR model.
 Sections \ref{sec:modeling} and \ref{sec:modeling2} are devoted to the theoretical analysis of learning the multi-agent PSR model of dynamical systems through a tensor decomposition.
 We present experimental study on several domains in Section \ref{sec:Experiments}.
 In Section \ref{sec:relatedwork}, we discuss related works on learning PSR.
 Finally, we conclude our work and give some suggestions on the future work.

 %{
 %\begin{itemize}
 %\item Difficulties:
 %(1). how to embed tensor into PSRs, get model prediction parameters and state vectors, and then its prediction equation.
 %(2). how to construct the auxiliary matrix from the obtained state factor matrix to solve the transition parameters of the model and obtain the state update equation. ( transformation matrix is hard to get)
 %
 %\item Techniques are based on TPSR model:
 %(1). we use two major tensor decomposition routes, CP and Tucker decomposition to extract the key information embedded in the tensor, and derive the prediction parameters and state vectors directly from the decomposition results. (the 3rd dimension of tensor  corresponds to state vectors, the other two dimensions correspond to prediction parameters)
 %(2). construct some auxiliary matrices, and obtain the transition parameters via linear regression (an optimization model)
 %\end{itemize}
 %}
 %\section{Preparations}\label{sec:preparations}
 \section{Technical Background of Multi-agent PSRs}\label{sec:preparations}
 Linear PSRs are a systematic-studied type of PSRs for modelling a dynamic system \cite{Littman2001Predictive}.
 %An agent interacts with dynamical environments by executing acts and receiving observations.
 The dynamic environment considered here is a discrete-time, controlled dynamic system with $N$-agent ($N\geq 2$), which produces a sequence of actions and observations with one action and one observation per time step.
 We extend all necessary definitions of a single-agent PSR model to a multi-agent PSR in this section.
 In order to clearly represent various notations, we use non-bold lowercase letters, boldface lowercase
 letters, capital letters, calligraphic letters and so on.
 We summarize a set of main notations in Table \ref{smb}.
 
 \begin{table}[h]
 	\centering
 	\caption{Summary of notations}
 	\vspace{0.5em}
 	\begin{tabular}{|c|c|}
 		\hline
 		$\textit{a,o,t,h},\phi$& action, observation, test, history, null (joint) history\\
 		$\textit{A,O,T,H}$& the set of actions, observations, tests and histories\\
 		$\text{Q},\textit{D}$& core test set, system dynamics matrix\\
 		$m_{t}$&  projection vector\\
 		$m_{ao},M_{ao}$&  (one-step) projection vector and transition matrix\\
 		\hline
 		$\textbf{a,o,t,h}$& joint action, joint observation, joint test,  joint history\\
 		$\mathcal{A,O,T,H}$& the set of joint actions, joint observations, joint tests and joint histories\\
 		$\textbf{Q},\textbf{H},\mathcal{Q}$& core joint test set, core joint history set, core test tensor\\
 		$\textbf{D},\mathcal{D}$& system dynamics matrix, system dynamics tensor\\
 		$m_{\textbf{t}},m_{{i_1}\dots {i_N}}$&  projection vector\\
 		$m_{\textbf{ao}},M_{\textbf{ao}}$&  (one-step) projection vector and transition matrix\\
 		$\mathcal {H}_{s}$&  joint history set at time step $s$\\
 		\hline
 		$\tilde{m}_{*},\tilde{M}_{*}$&  projection vector and transition matrix of TPSR\\
 		$ F$ & projection matrix of TPSR\\
 		\hline
 		{$s$}& time step\\
 		$\Phi_d,\Psi_d$& training dataset and test dataset\\ 	
 		$p(\cdot),p(\cdot|\cdot)$& probabilistic operator and conditional probabilistic operator\\
 		$\mathcal{I}(e,\mathcal{S})$& an index that records the index of the element $e$ in set $\mathcal{S}$\\
 		$\mathcal{I}(\mathcal{S}^\prime,\mathcal{S})$& an index set that records the indices of elements of subset $\mathcal{S}^\prime$ in set $\mathcal{S}$\\
 		\hline
 		$A,A_{i:},A_{:j}$& a matrix, its $i$-th row vector and $j$-th column vector\\
 		$A ^{T}, A^{-1}$& the transpose and (pseudo-)inverse of matrix $A$\\
 		$A_{\mathcal{I}}$& a sub-matrix of $A$ consisting of rows indicated by  set $\mathcal{I}$  \\
 		%		$I$& a row or column index table of matrix\\
 		%		$A_{I:},A_{:I}$& a matrix consisting of $I$ from the matrix $A$\\		
 		%		$sgn(\cdot)$&Sign function\\
 		$A^{(i)}$&  the $i$-th factor matrix in Tucker decomposition \\	
 		\hline
 		$\mathcal{A}$& a tensor\\
 		%		$\mathcal{A},\mathcal{A}^{(i)}$& any tensor and $i$-th auxiliary tensor \\
 		$\mathcal{A}_{(k)}$&  mode-$k$ matricization of tensor $\mathcal{A}$\\
 		\hline
 		%		$\mathcal{A}_{(i::)},\mathcal{A}_{(:j:)},\mathcal{A}_{(::k)}$& $i$-th horizontal, $j$-th lateral, and $k$-th frontal slide of a tensor $\mathcal{A}$\\
 		$\times_{n}$& the $n$-mode product of a tensor with a matrix\\
 		$*,\circ, \otimes$& Hadamard product, outer product, Kronecker product of vectors\\
 		\hline	
 	\end{tabular}
 	\label{smb}
 \end{table}
 
 \subsection{Single-agent PSR}
 In a controllable dynamical system with a single agent, the agent continuously performs a sequence of actions $\textit{a}$ chose from the action set $A=\{a^1,a^2,\dots,a^{|A|}\}$ according to any policy $\pi$ and senses a sequence of observations $\textit{o}$ which can be identified in the observation set $O=\{o^1,o^2,\dots,o^{|O|}\}$. At time step $s$, the agent has already experienced a sequence of action-observation pairs that are called as {\em history} $h$, i.e. $h_s=\textit{a}_1 \textit{o}_1\dots \textit{a}_{s} \textit{o}_{s}$ for any  possible history at time step $s$. All possible histories at the entire horizon forms a {\em history set} $H$. After an agent applies $ao$ at time step $s$, the history $h_s$ is updated to history $h_{s+1}=h_sao$. The agent may expect to follow a special sequence of action-observation pairs with the length $l$ which is called as {\em test} $t$ beginning immediately at time step $s$, i.e. $t=\textit{a}_1 \textit{o}_1\dots \textit{a}_{l} \textit{o}_{l}$. All possible tests in the future form a {\em test set} $T$. The probability of the occurrence of a test $t$ given the history $h_s$ is denoted as $p(t|h_s)$, which can be calculated by {\em prediction equation}, thus
 $$p(t|h_s)=f(p(\text{Q}|h_s))=p(\text{Q}|h_s)m_{t}$$
 where $f$ is the project function in a linear PSR,  $p(\text{Q}|h_s)$ is the state vector at time step $s$, and $m_{t}$ is the projection vector of test $t$. The new state vector of a linear PSR is calculated by {\em updating equation}, thus $$p(\text{Q}|{h}_{s+1})=\frac{p(\text{Q}|{h}_s) M_{{ao}}}{p(\text{Q}|{h}_s)\bm_{{ao}}}$$
 where $m_{ao}$ is the projection vector for $t=ao$ and $M_{ao}$ is a matrix consisted of one-step extension projection vectors. In short, a single agent linear PSR model in a controlled partially observable system has the parameters $<{A}, {O}, \text{Q}, \{\bm_{{ao}}\},\{M_{{ao}}\}, p(\text{Q}|\phi)>$:
 the set of actions ${A}$, the set of  observations ${O}$, the set of core tests $\text{Q}$,
 the model parameters $\bm_{{ao}}$ and $M_{{ao}}$ ($\forall~{a}\in {A}, {o}\in{O}$), and
 an initial prediction vector $p(\text{Q}|\phi)$, where $\phi$ is the null history at initial time step $s=0$.
 
 \subsection{Multi-agent PSR}
 For a dynamic $N$-agent system,  the {\em joint action} $\textbf{a}= a^{(1)}a^{(2)}\cdots a^{(N)}$ represents a sequence of executable actions that agents, e.g. agent (1), $\cdots$, ($N$), can operate simultaneously, and  the {\em joint observation} $\textbf{o}= o^{(1)}o^{(2)}\cdots o^{(N)}$ represents the observations that the agents may receive in their interactions.
 % where $a^{(i)}$ and $o^{(i)}$  represent that agent $i$ performs the action $a^{(i)}$ and receives the observation $o^{(i)}$ in the interactive progress, respectively.
 Then we have the action set $A^p=\{{a}^1,{a}^2,\ldots,{a}^{|{A}|-1}, {a}^{|{A}|}\}$ to represent all valid actions that  agent $p$ can perform, and the observation set $O^p=\{{o}^1,{o}^2,\ldots,{o}^{|{O}|-1}, {o}^{|{O}|}\}$ for all observations that the agent $p$ may receive in the interaction.
 Hence, we assume that $\mathcal{A}=\{\textbf{a}^1,\textbf{a}^2,\ldots,\textbf{a}^{|\mathcal{A}|-1},\textbf{a}^{|\mathcal{A}|}\}$ is a set of all executable joint actions that agents can operate and $\mathcal{O}=\{\textbf{o}^1,\textbf{o}^2,\ldots,\textbf{o}^{|\mathcal{O}|-1},\textbf{o}^{|\mathcal{O}|}\}$ is a set of all joint observations that the agents may receive.
 
 A {\em joint test} $\textbf{t}=\textbf{a}_1\textbf{o}_1\textbf{a}_2\textbf{o}_2\cdots \textbf{a}_l\textbf{o}_l$ represents the sequence of joint action-observation pairs of all the agents that they may encounter in the future.
 Accordingly, we have the action sequence $\textbf{t}^\textbf{a}=\textbf{a}_1\textbf{a}_2\cdots \textbf{a}_l$ and the observation sequence $\textbf{t}^\textbf{o}=\textbf{o}_1\textbf{o}_2\cdots \textbf{o}_l$. The {\em joint test set} is a set of all possible joint tests $\textbf{t}$ of the agents, denoted as $\mathcal{T}$. A {\em test} is a limited sequence of action-observation pairs in a single agent scenario, e.g., test $t_{i_1}^{(i)}=a_1^{(i)}o_1^{(i)} a_2^{(i)}o_2^{(i)}\cdots a_l^{(i)}o_l^{(i)}$ represents the ${i_1}$-th sequence of action-observation pairs of the $i$-th agent's test set $\mathcal{T}^{(i)}$. Then a joint test can be expressed by using the test of all the agents, e.g., $\textbf{t}_{i}= t_{i_1}^{(1)}\cdots t_{i_N}^{(N)}$,
 %for simplicity $\textbf{t}^{i}= t_{i_1}\dots t_{i_N}$,
 which is the $i$-th joint test in joint test set $\mathcal{T}$.
 A {\em joint history} has the same structure as the joint test, which is used to describe the entire sequence of past action-observation pairs, e.g.,
 joint history $\textbf{h}_s=\textbf{a}_1\textbf{o}_1\textbf{a}_2\textbf{o}_2\ldots \textbf{a}_s\textbf{o}_s$ at time step $s$ {and the joint history will be updated to $\textbf{h}_{s+1}=\textbf{h}_{s}\textbf{ao}$ after agents taking the joint action $\textbf{a}$ and seeing the joint observation $\textbf{o}$ from the joint history $\textbf{h}_s$.
 	The {\em joint history set} is a set of all possible joint histories $ \textbf {h} $ of the agents, denoted as $ \mathcal {H}$.} And we denote a joint history set of all possible joint histories $ \textbf {h}_s $ of the agents with length $s$, denoted as $ \mathcal {H}_{s}\subset \mathcal {H}$, i.e., $ \mathcal {H}_{s}=\{\textbf{h}|\textbf{h}\in \mathcal{H},|\textbf{h}|=s\}$, which contains all possible joint histories that agents have encountered at time step $s$. Therefore, the  joint history set $\mathcal{H}$ can be described as $\mathcal{H}=\{\mathcal {H}_{s=0}=\phi,\mathcal {H}_{s=1},\dots,\mathcal {H}_{s=L}\}$ sampled from training dataset $\Phi_d$, where $L$ is max-length of sequences of action-observation in $\Phi_d$.
 
 A {\em sequence prediction} problem is defined as predicting the probabilities of different joint observation sequences when agents execute the joint action sequence given an arbitrary history. Thus, to make a prediction of a joint test $\textbf{t}$ given the prior joint history $\textbf{h}_s\in \mathcal{H}_{s}$ at time step $s$, denoted by $p(\textbf{t}|\textbf{h}_s)$,
 is defined as
 \begin{equation}\label{pth}
 p(\textbf{t}|\textbf{h}_s)= prob(\textbf{t}^\textbf{o}|\textbf{h}_s \textbf{t}^\textbf{a}).
 \end{equation}
 
 For any set of joint tests $\textbf{Q}=\{\textbf{q}_i|\textbf{q}_i\in \mathcal{T}, i\in \{1,2,\dots,n\}\}$,
 its prediction vector (or state vector) is $p(\textbf{Q}|\textbf{h}_s)=[p(\textbf{q}_1|\textbf{h}_s), \ldots, p(\textbf{q}_n|\textbf{h}_s)] \in \mathbb{R}^{1\times n}$.
 If $p(\textbf{Q}|\textbf{h}_s)$ forms {\em sufficient statistic} at any joint history $\textbf{h}_s$ in the dynamic system at time step $s$, %{which represents the belief state of the dynamic system. That is,}
 i.e., all tests can be predicted based on $p(\textbf{Q}|\textbf{h}_s)$~(in other words, there exists a function $f_\textbf{t}$ such that
 $p(\textbf{t}|\textbf{h}_s) = f_\textbf{t}(p(\textbf{Q}|\textbf{h}_s))$ for any test $\textbf{t}$), then the set $\textbf{Q}$ is called {\em  core joint test set}.
 We will denote $p(\textbf{Q}|\textbf{h}_{s}\leftarrow\textbf{h}_k)$ as $p(\textbf{Q}|\ \textbf{h}_k)$ for any given joint history $\textbf{h}_k\in \mathcal{H}$($k\in[1,|\mathcal{H}|$) at time step $s=|\textbf{h}_k|$ for simplicity.
 For linear PSRs, the function $f_\textbf{t}$ is a linear function.
 Thus, the prediction formula Eq.~(\ref{pth}) can be rewritten as
 \begin{equation}
 p(\textbf{t}|\textbf{h}_s)=p(\textbf{Q}|\textbf{h}_s) \bm_\textbf{t} \label{predictive},
 \end{equation}
 where $\bm_\textbf{t}\in \mathbb{R}^{n\times 1}$ is called {\em projection vector}. {For $\textbf{t}_{i}= t_{i_1}^{(1)}\cdots t_{i_N}^{(N)}$, Eq.~(\ref{predictive}) becomes
 	\begin{equation}
 	p(t_{i_1}^{(1)}\cdots t_{i_N}^{(N)}|\textbf{h}_s)=p(\textbf{Q}|\textbf{h}_s) \bm_{t_{i_1},\dots ,t_{i_N}} \label{predictiven},\nonumber
 	\end{equation}
 	we will denote $\bm_{t_{i_1},\dots ,t_{i_N}}$ as $\bm_{{i_1}\dots {i_N}}$ for simplicity. For example, in a 2-agent system, the projection vector for a joint test $\textbf{t} =t_i^{(1)}t_j^{(2)}$ can be denoted by $\bm_{ij}$. The projection vector for a one-step joint test $\textbf{t}=\textbf{ao}$ in a multi-agent system is denoted by
 	$m_{\textbf{ao}}$.
 }
 %we define a special projection vector $
 %m_{\textbf{ao}}$ corresponding to a one-step joint test $\textbf{t}$, i.e.,  $\textbf{t}=\textbf{ao}=a_{i_1}^{(1)}  o_{i_1}^{(1)}a_{i_2}^{(2)} o_{i_2}^{(2)}\dots a_{i_N}^{(N)} o_{i_N}^{(N)}$, which can be obtained through finding a $\bm_{{i_1}\dots {i_N}}$ has subscript ${i_1}\dots {i_N}$.
 %}
 
 When a system receives the agents' joint action $\textbf{a}$, it immediately transforms into a next state, which means the PSR model should update its state at the same time.
 The update calculates the new state $p(\textbf{Q}|\textbf{h}_{s+1})$ from the previous state $p(\textbf{Q}|\textbf{h}_s)$ after agents take the joint action $\textbf{a}$ and receive the observation $\textbf{o}$ from the history $\textbf{h}_s$. The initial state is $p(\textbf{Q}|\phi)$ when $\textbf{h}_s$ takes the null joint history $\phi$ at time step $s=0$.
 %Need to mention that the initial state of the dynamic system is $p(\textbf{Q}|\phi)$, which means the joint history is initialized as empty joint history with the beginning of the system, i.e., $\textbf{h}_1\leftarrow \phi$.
 For any core joint test $ \textbf{q}_i \in  \textbf{Q}$ and $\textbf{h}_{s+1}=\textbf{h}_{s}\textbf{ao}$, $\forall\,  \textbf{a}\in \mathcal{A}, \textbf{o}\in \mathcal{O}$, we compute the update as follows:
 \begin{equation}
 p(\textbf{q}_i|\textbf{h}_{s+1})=\frac{p(\textbf{ao}\textbf{q}_i|\textbf{h}_s)}{p(\textbf{ao}|\textbf{h}_s)}=\frac{p(\textbf{Q}|\textbf{h}_s)\bm_{\textbf{ao}\textbf{q}_i}}{p(\textbf{Q}|\textbf{h}_s)\bm_{\textbf{ao}}},  \label{update}
 \end{equation}
 where $\bm_{\textbf{ao}}$ and  $\bm_{\textbf{aoq}_i}$ are $\bm_\textbf{t}$ for each one-step joint test ($\textbf{ao}$) and each one-step extension ($\textbf{ao}\textbf{q}_i$) respectively. The first equality of Eq.~(\ref{update}) is obtained by Bayes rule, and the second one is computed by Eq.~(\ref{predictive}). By defining the matrix $M_{\textbf{ao}}\in \mathbb{R}^{n\times n}$,
 in which the $i$-th column vector is $\bm_{\textbf{ao}\textbf{q}_i}$, we have
 $$M_{\textbf{ao}} = [ \bm_{\textbf{ao}\textbf{q}_1} ~~ \bm_{\textbf{ao}\textbf{q}_2}~~ \cdots~~ \bm_{\textbf{ao}\textbf{q}_n}], ~~\forall\, \textbf{a}\in \mathcal{A}, \textbf{o}\in \mathcal{O}.$$
 %\end{flalign}
 Subsequently, Eq.~(\ref{update}) can be rewritten as
 \begin{equation}\label{updaten}
 p(\textbf{Q}|\textbf{h}_{s+1})=\frac{p(\textbf{Q}|\textbf{h}_s) M_{\textbf{ao}}}{p(\textbf{Q}|\textbf{h}_s)\bm_{\textbf{ao}}}.
 \end{equation}
 %where the $i$-th column of $M_{\textbf{ao}}$ is  $\bm_{\textbf{ao}\textbf{q}_i}, \forall \textbf{a}\in \mathcal{A}, \textbf{o}\in \mathcal{O}$.
 The vectors \{$\bm_{\textbf{ao}}$\} and matrices $\{M_{\textbf{ao}}\}$ ($\forall \textbf{a}\in \mathcal{A}, \textbf{o}\in \mathcal{O})$  are called the model parameters of the linear PSR model.
 {%It can be seen that the prediction vector of $\textbf{h}_k$ at any time of the system can be obtained step by step from the prediction vector $p(\textbf{Q}|\phi)$ at the time of the empty history $\phi$.
 	If an initial prediction vector $p(\textbf{Q}|\phi)$ for a given null joint history $\phi$, the prediction vector $p(\textbf{Q}|\textbf{h}_s)$ can be calculated step by step for any time $s$.
 	In addition, for any test $\textbf{t} = \textbf{a}_1\textbf{o}_1\textbf{a}_2\textbf{o}_2\cdots \textbf{a}_l\textbf{o}_l$, its corresponding projection vector $m_\textbf{t}$ can be computed by the chain rule in terms of conditional probability, Eqs.~(\ref{pth}) and~(\ref{updaten}), i.e.,
 	\begin{flalign}\label{eq:36mt}
 	m_\textbf{t}= M_{\textbf{a}_1\textbf{o}_1} M_{\textbf{a}_2\textbf{o}_2}\cdots M_{\textbf{a}_{l-1}\textbf{o}_{l-1}}m_{\textbf{a}_{l}\textbf{o}_{l}}.\nonumber
 	\end{flalign}
 	Notice that when the model parameters and the initial prediction vector of the linear PSR model are known, i.e., the modelling of the entire PSR model is completed, we can make the sequential prediction $p(\textbf{t}|\textbf{h})$.}
 If the core {joint} tests $\textbf{Q}$ is found, the parameters can be computed as follows.
 \begin{equation}\label{updateparameters}
 \begin{array}{l}
 % \nonumber to remove numbering (before each equation)
 \bm_{\textbf{ao}} = p(\textbf{Q}|\textbf{H})^{-1} p(\textbf{ao}|\textbf{H}),  \nonumber\\
 \bm_{\textbf{aoq}_i}=   p(\textbf{Q}|\textbf{H})^{-1} p(\textbf{aoq}_i|\textbf{H}),\nonumber
 \end{array}\nonumber
 \end{equation}
 where $\textbf{H}$ is {called \textit{core joint history set}. }%, and mathematical operators $\textbf{}^{-1}$ is a (pseudo)~inverse of the matrix.
 The values of $p(\textbf{Q}|\textbf{H}),$ $p(\textbf{ao}|\textbf{H})$ and $p(\textbf{aoq}_i|\textbf{H}) $ can be estimated through the training data in the following.
 \begin{flalign}%\label{eq:310PQH}
 p(\textbf{Q}|\textbf{H}) =
 \begin{bmatrix}
 p(\textbf{q}_1|\textbf{h}_1) & p(\textbf{q}_2|\textbf{h}_1) &\cdots &p(\textbf{q}_n|\textbf{h}_1)  \\ p(\textbf{q}_1|\textbf{h}_2) & p(\textbf{q}_2|\textbf{h}_2) &\cdots &p(\textbf{q}_n|\textbf{h}_2) \\
 \vdots & \vdots &\ddots&\vdots \\
 p(\textbf{q}_1|\textbf{h}_{|\textbf{H}|}) & p(\textbf{q}_2|\textbf{h}_{|\textbf{H}|}) &\cdots &p(\textbf{q}_n|\textbf{h}_{|\textbf{H}|})
 \end{bmatrix},\nonumber
 \end{flalign}
 \begin{flalign}\label{eq:311PaoH}
 p(\textbf{ao}|\textbf{H}) =
 \begin{bmatrix}
 p(\textbf{ao}|\textbf{h}_1) \\
 p(\textbf{ao}|\textbf{h}_2) \\
 \vdots \\
 p(\textbf{ao}|\textbf{h}_{|\textbf{H}|})
 \end{bmatrix},~
 \text{and} \quad
 %\end{flalign}
 %
 %\begin{flalign}\label{eq:312PaoqiH}
 p(\textbf{aoq}_i|\textbf{H}) =
 \begin{bmatrix}
 p(\textbf{aoq}_i|\textbf{h}_1) \\
 p(\textbf{aoq}_i|\textbf{h}_2) \\
 \vdots \\
 p(\textbf{aoq}_i|\textbf{h}_{|\textbf{H}|})
 \end{bmatrix}.
 \nonumber
 \end{flalign}
 
 In a summary, a linear PSR model in a controlled partially observable system has the parameters $<\mathcal{A}, \mathcal{O}, \textbf{Q}, \{\bm_{\textbf{ao}}\},\{M_{\textbf{ao}}\}, p(\textbf{Q}|\phi)>$:
 the set of joint actions $\mathcal{A}$, the set of joint observations $\mathcal{O}$, the set of core {joint} tests $\textbf{Q}$,
 the model parameters $\bm_{\textbf{ao}}$ and $M_{\textbf{ao}}$ ($\forall~\textbf{a}\in \mathcal{A}, \textbf{o}\in \mathcal{O}$), and
 an initial prediction vector $p(\textbf{Q}|\phi)$, where $\phi$ is the null joint history.
 The process of finding $\textbf{Q}$ is called  {\em the discovery problem}, while the computation of the projection vectors by using $\textbf{Q}$ to represent all the other tests is called  {\em the learning problem}.
 
 In a variant of PSRs, transformed predictive representation~(TPSR) \cite{Rosencrantz2004Learning} tries to maintain a small number of linear combinations of the
 probabilities of a larger number of tests instead of maintaining probability distributions over the outcomes of a small set of tests. {Therefore, we are aiming to learn a multi-agent PSR model based on TPSR in this paper. Traditionally,  the parameters $<\mathcal{A}, \mathcal{O},x_s, \{\tilde{m}_{\textbf{ao}}\},\{\tilde{M}_{\textbf{ao}}\},x_0>$ describes a TPSR model,
 	where $x_s$ is the compressed state vector, $x_0$ is an initial compressed state vector and the other parameters are the same as a usual linear PSR model.
 	%	the set of joint actions $\mathcal{A}$, the set of joint observations $\mathcal{O}$, compressed state vector $x_k$,
 	%	the set of projection vectors $\{\tilde{m}_{\textbf{ao}}\}$, the set of transition matrixes $\{\tilde{M}_{\textbf{ao}}\}$, and
 	%	an initial compressed state vector $x_1$, where $x_1$ is the compressed state vector of the initial state of the system with null history.
 	%By defining a compressed state vector $x_k\in \mathbb {R} ^ {R}$,
 	In fact, $x_s$ is a compressed version of system prediction vector $p (\textbf{Q}|\textbf{h}_s)$, and hence $x_{0}$ is a reduced version of an initial prediction vector $p (\textbf{Q}|\textbf{h}_{s=0}=\phi)$.}
 Therefore, $x_s\in \mathbb {R} ^ {1\times R}$  can be calculated by multiplying $p (\textbf{Q}|\textbf{h}_s)$ by a projection matrix ${F}$, namely:
 \begin{equation}
 x_s = p(\textbf{Q}|\textbf{h}_s){F},
 \label{eq:631}
 \end{equation}
 where the matrix ${F}\in \mathbb{\R}^{|\textbf{Q}| \times R}$, $|\textbf{Q}|$ is the size of core joint test set $\textbf{Q}$.
 Therefore, for a given joint history $\textbf{h}_k\in \mathcal{H}$($k\in[1,|\mathcal{H}|]$) at time step $s=|\textbf{h}_k|$, we have $x_s= p(\textbf{Q}|\textbf{h}_k){F}$ and can denote $x_s$ as $x_k$ for simplicity.
 %\textcolor{blue}{, the subscript $k$ of $x_k$ corresponds to $\textbf{h}_k$ one by one., and $R$ is a small real number.}
 Hence, for a given joint history $\textbf{h}_k\in \mathcal{H}$($k\in[1,|\mathcal{H}|]$) at time step $s=|\textbf{h}_k|$, the prediction formula Eq.~(\ref{predictive}) and the state update Eq.~(\ref{updaten}) can be rewritten as
 \begin{equation}
 p(\textbf{t}|\textbf{h}_k)=x_k \tilde{m}_\textbf{t} \label{predictivetpsr},
 \end{equation}
 and
 \begin{flalign}\label{eq:661}
 x_{k+1}=\frac{x_{k} \tilde{M}_{\textbf{ao}}}{x_{k} \tilde{m}_{\textbf{ao}}},
 \end{flalign}
 {where $\tilde{m}_{\textbf{ao}} \in \mathbb {R} ^ {R\times 1}$ is called a projection vector or model prediction parameter, which can be calculated in Eqs.~(\ref{predictive}), (\ref{eq:631}) and (\ref{predictivetpsr}), i.e., $\tilde{m}_{\textbf{ao}}= {F}^{-1}m_{\textbf{ao}}$ , and $\tilde{M}_{\textbf{ao}}\in \mathbb {R} ^ {R\times R}$ is called a transition matrix or model update parameter, which can be calculated in Eqs.~(\ref{updaten}), (\ref{eq:631}), (\ref{eq:661}) and $\tilde{m}_{\textbf{ao}}= {F}^{-1}m_{\textbf{ao}}$, i.e., $\tilde{M}_{\textbf{ao}}= {F}^{-1}M_{\textbf{ao}}{F}$.}
 However, it is difficult to solve the transformation
 matrix ${F}$ directly in order to learn the model parameters.
 Generally, the two parameters learned for TPSR will not be identical to those learned in a traditional PSR model.
 It can no longer interpret the elements of $x_{k}$ as probabilities as they may be negative or even larger than 1.
 {Since the two parameters fully summary the system updating rule and sequential prediction, we will
 	learn them in another way without the help of the projection matrix ${F}$ in this paper.}
 %Need to mention that it is not necessary to explicitly give the projection matrix ${F}$ and core joint test set $\textbf{Q}$, because the transition matrix and prediction parameter can fully summary the system updating progress and sequential prediction. Therefore, it is not necessary to solve the transformation matrix ${F}$ directly in order to learn the model parameters if we can learn these parameters in the other way. }
 \section{System Dynamics Tensor for Multi-agent PSRs}\label{sec:SDT}
 {
 	In this section, we construct the system dynamics matrix as a high order tensor for learning a multi-agent PSR model, and give the formulas to calculate the system marginal dynamic matrix and the PSR model under each agent's perspective. The PSR model of each agent can be used to predict its future sequences given the experienced history of the system.}
 
 {
 	We propose a {\em system dynamics tensor} for learning a multi-agent PSR model based on a tensor approach.
 	For a system with $N$ agents, we use $\mathcal{D}\in \mathbb{\R}^{n_1\times n_2 \times n_3 \times \cdots \times  n_{N+1}}$ to denote a system dynamics tensor, whose element is $\mathcal{D}_{i_1 i_2\cdots i_N k} =p(t_{i_1}^{(1)}\dots t_{i_N}^{(N)}|\textbf{h}_k)$, which can be estimated by the reset algorithm \cite{James2004Learning}, where $t_{i_1}^{(1)}$, $t_{i_2}^{(2)}$, $\ldots$, $t_{i_N}^{(N)}$ are the tests for each agent and $\textbf{h}_k\in \mathcal{H}$($k\in[1,|\mathcal{H}|)$) is the given joint history at time step $s=|\textbf{h}_k|$. Then,  {\em the discovery problem} is transferred into finding a minimal linearly independent set (i.e., core joint test set \textbf{Q}),
 	%among mode-$(N+1)$ fibers of tensor $\mathcal{D}$,
 	so that the whole fibers listed in the set \textbf{Q} form a basis of the space spanned by the mode-$(N+1)$ fibers of tensor $\mathcal{D}$,
 	%so that all fibers are a weighted sum of those fibers.
 	and these fibers together form a sub-tensor,
 	%of  $\mathcal{D}$,
 	namely
 	%we call the tensor constructed by mode-3 fibers extracted from the system dynamics tensor $\mathcal{D}$ as
 	\textit{core test tensor} $\mathcal{Q}$.}
 
 {
 	Without loss of generality, let us discuss a 2-agent scenario, the system dynamic tensor
 	$\mathcal{D}\in \mathbb{\R}^{n_1\times n_2 \times n_3} $ is a 3rd order tensor now,
 	where $n_{3}=|\mathcal{H}|$ and $n_i=|\mathcal{T}^{(i)}|(i=1, 2)$,
 	%whose element is $\mathcal{D}_{ijk} = p(t_i^{(1)}t_j^{(2)}|\textbf{h}_k)$,
 	as intuitively shown in Fig. \ref{Fig.smdtij}.
 	Its element, e.g., $p(t_1^{(1)}t_2^{(2)}|{\textbf{a}^1\textbf{o}^1})$, is corresponding to the test $t_1^{(1)}={a^1 o^2}$ in test set $\mathcal{T}^{(1)}$ of agent 1, the test $t_2^{(2)}={a^2 o^2}$ in test set $\mathcal{T}^{(2)}$ of agent 2, and the joint history ${\textbf{a}^1\textbf{o}^1}$ in joint history set $\mathcal{H}$.
 	The corresponding \textit{system marginal dynamics matrices} (i.e., $D^{(1)}\in \mathbb{\R}^{|\mathcal{T}^{(1)}|\times|\mathcal{H}^{(1)}|} $ and $D^{(2)}\in \mathbb{\R}^{|\mathcal{T}^{(2)}|\times|\mathcal{H}^{(2)}|}$) can be calculated by the elements of system dynamics tensor $\mathcal{D}$ according to the probability theory.
 	Let us take the computation of $D^{(1)}$ as an example.
 	For any $\textbf{t}^{(1)}\in\mathcal{T}^{(1)}$, $ \textbf{h}^{(1)}\in\mathcal{H}^{(1)}$,
 	the element $D^{(1)}_{\textbf{t}^{(1)}, \textbf{h}^{(1)}}$ is computed as follows:
 	\begin{eqnarray*}
 		% \nonumber to remove numbering (before each equation)
 		D^{(1)}_{\textbf{t}^{(1)},\textbf{h}^{(1)}} &=& {p(\textbf{t}^{(1)}|\textbf{h}^{(1)})}  \\
 		&=& \sum_{\textbf{h}^{(2)},|\textbf{h}^{(2)}|=|\textbf{h}^{(1)}|} ~~{\sum_{\textbf{t}^{(2)},|\textbf{t}^{(2)}|=|\textbf{t}^{(1)}|} {p(\textbf{t}^{(1)}\textbf{t}^{(2)}|\textbf{h}^{(1)}\textbf{h}^{(2)})}} \\
 		&=& \sum_{\textbf{h}^{(2)},|\textbf{h}^{(2)}|=|\textbf{h}^{(1)}|} ~~{\sum_{\textbf{t}^{(2)},|\textbf{t}^{(2)}|=|\textbf{t}^{(1)}|} {\mathcal{D}_{\textbf{t}^{(1)}\textbf{t}^{(2)},\textbf{h}^{(1)}\textbf{h}^{(2)}}}},
 	\end{eqnarray*}
 	where the second equation is obtained by summing up all the histories and tests of all the other agents (i.e., agent 2 in this case) with length $|\textbf{h}^{(2)}|=|\textbf{h}^{(1)}|$ and $|\textbf{t}^{(2)}|=|\textbf{t}^{(1)}|$, marked as two
 	red directions on the upper left hand side of Fig. \ref{Fig.smdtij}. For simplicity, we write the two system marginal dynamics matrices as follows.
 	%
 	%To be specific, after merging the same subsequences in the other two dimensions, we can easily obtain the system marginal dynamics matrices $D^{(1)}$ and $D^{(2)}$ as follows:
 	%
 	%$\forall \textbf{h}^{(i)}\in\mathcal{H}^{(i)}$,$\textbf{t}^{(i)}\in\mathcal{T}^{(i)}$, the element $D^{(i)}_{\textbf{h}^{(i)},\textbf{t}^{(i)}}$ of agent $i$'s system marginal dynamics matrix can be calculated from system dynamics tensor $\mathcal{D}$, through
 	%We drop the constraints in Eq.~(\ref{smdtij}) for simplicity.
 	%}
 	\begin{equation}\label{smdtij}
 	\begin{aligned}
 	D^{(1)}=\sum_{\textbf{h}^{(2)}\in\mathcal{H}^{(2)}} {\sum_{\textbf{t}^{(2)}\in\mathcal{T}^{(2)}} \mathcal{D}}
 	,~\text{and} \quad
 	D^{(2)}=\sum_{\textbf{h}^{(1)}\in\mathcal{H}^{(1)}} {\sum_{\textbf{t}^{(1)}\in\mathcal{T}^{(1)}} \mathcal{D}}
 	\end{aligned}
 	.\nonumber
 	\end{equation}
 	The two dimensions of each matrix reflect test and history, respectively.
 	%\begin{equation}\label{smdtij}
 	%\begin{aligned}
 	%D^{(1)}=\sum_{\textbf{h}^{(2)}} {\sum_{\textbf{t}^{(2)}} \mathcal{D}}
 	%,~\text{and} \quad
 	%D^{(2)}=\sum_{\textbf{h}^{(1)}} {\sum_{\textbf{t}^{(1)}} \mathcal{D}}
 	%\end{aligned}
 	%.\nonumber
 	%\end{equation}
 	\begin{figure}[!h]
 		\centering
 		\includegraphics[height=11cm, width=16cm,keepaspectratio]{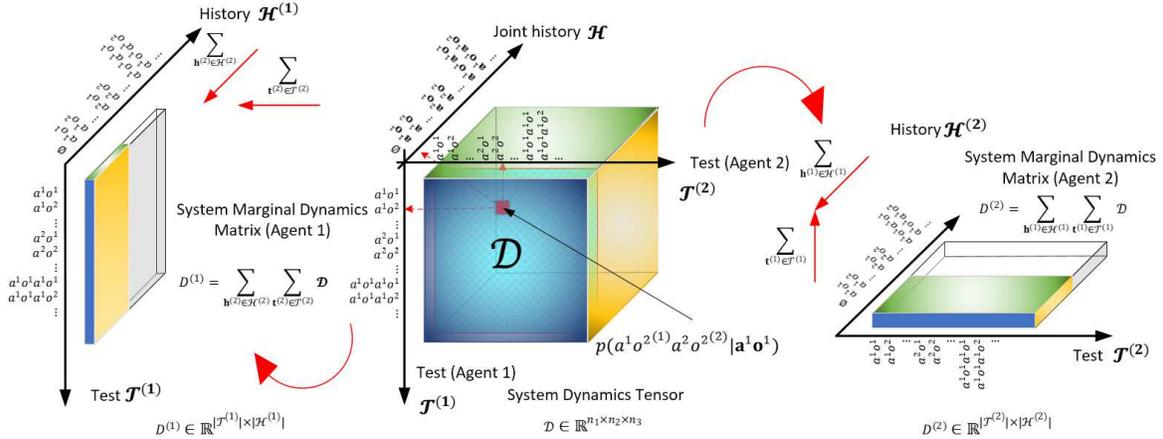}
 		\caption{Diagram of system marginal dynamics matrices obtained by system dynamics tensor of 2-agent system}
 		\label{Fig.smdtij}
 	\end{figure}
 	
 	%{
 	%%As shown in Fig. \ref{Fig.smdtij}, the system dynamics tensor of two agents system can be depicted as a 3-dimensional tensor, whose coordinates are respectively corresponding to the test set of the $1$-st agent, the test set of the $2$-nd agent and the joint history set of the 2-agent system, and can index an unique sequence in the each set.
 	%After merging the same subsequences in the other two dimensions, we can easily obtain the system marginal dynamics matrix, which is a matrix of history and test in single agent.}

 	%{
 	On the other hand, the {\em system dynamics matrix} \cite{Littman2001Predictive} can be employed for learning a multi-agent PSR model as an alternative way since we will apply the traditional matrix-based single-agent PSR learning algorithms
 	to the multi-agent PSRs in Section \ref{sec:comparativemethods}.
 	The matrix consists of joint histories and tests, and their elements can be estimated by the reset algorithm \cite{James2004Learning}. The difference between system dynamics matrix and  system dynamics tensor  is that we put the joint tests of all the agents in only one dimension when constructing a system dynamics matrix.
 	%We can simply apply the traditional matrix-based single-agent PSR learning algorithms to multi-agent PSR by simply redefining the {\em system dynamics matrix} \cite{Littman2001Predictive} .
 	
 	For a two-agent system, the system dynamics matrix $\textbf{D}\in \mathbb{\R}^{|\mathcal{T}|\times|\mathcal{H}|}$ can be depicted as a two-dimensional matrix in Fig. \ref{Fig.smdij}. For example, the element $p({\textbf{a}^2\textbf{o}^1}|{\textbf{a}^1\textbf{o}^1})$ of matrix $\textbf{D}$ is corresponding to the joint test $\textbf{a}^2\textbf{o}^1$ in the joint test set $\mathcal{T}$, and the joint history ${\textbf{a}^1\textbf{o}^1}$ in the joint history set $\mathcal{H}$. The corresponding system marginal dynamics matrices (i.e., $D^{(1)}\in \mathbb{\R}^{|\mathcal{T}^{(1)}|\times|\mathcal{H}^{(1)}|} $ and $D^{(2)}\in \mathbb{\R}^{|\mathcal{T}^{(2)}|\times|\mathcal{H}^{(2)}|}$) can be obtained by the elements of matrix $\textbf{D}$.
 	Let us take the calculation of $D^{(1)}$ as an example. For any $\textbf{t}^{(1)}\in\mathcal{T}^{(1)}$, $ \textbf{h}^{(1)}\in\mathcal{H}^{(1)}$, the element $D^{(1)}_{\textbf{t}^{(1)},\textbf{h}^{(1)}}$ is calculated below.
 	\begin{eqnarray*}
 		% \nonumber to remove numbering (before each equation)
 		D^{(1)}_{\textbf{t}^{(1)},\textbf{h}^{(1)}}&=& p(\textbf{t}^{(1)}|\textbf{h}^{(1)}) \\
 		&=& \sum_{\textbf{h}^{(2)},|\textbf{h}^{(2)}|=|\textbf{h}^{(1)}|} ~~{\sum_{\textbf{t}^{(2)},|\textbf{t}^{(2)}|=|\textbf{t}^{(1)}|} {p(\textbf{t}^{(1)}\textbf{t}^{(2)}|\textbf{h}^{(1)}\textbf{h}^{(2)})}} \\
 		&=&\sum_{\textbf{h}^{(2)},|\textbf{h}^{(2)}|=|\textbf{h}^{(1)}|} ~~{\sum_{\textbf{t}^{(2)},|\textbf{t}^{(2)}|=|\textbf{t}^{(1)}|} {\textbf{D}_{\textbf{t}^{(1)}\textbf{t}^{(2)},\textbf{h}^{(1)}\textbf{h}^{(2)}}}},
 	\end{eqnarray*}
 	where the second equation is obtained by summing up all the histories and testes of all the other agents (i.e., agent 2 in this case) with length $|\textbf{h}^{(2)}|=|\textbf{h}^{(1)}|$ and $|\textbf{t}^{(2)}|=|\textbf{t}^{(1)}|$, as shown in the left hand side of Fig. \ref{Fig.smdij}. Hence, the two dimensions of $D^{(1)}$ reflect history and test of agent 1, respectively. For simplicity, we write the two system marginal dynamics matrices as follows.
 	\begin{equation*}\label{smdij}
 	\begin{aligned}
 	D^{(1)}=\sum_{\textbf{h}^{(2)}\in\mathcal{H}^{(2)}} {\sum_{\textbf{t}^{(2)}\in\mathcal{T}^{(2)}} \textbf{D}}, ~\text{and} \quad
 	D^{(2)}=\sum_{\textbf{h}^{(1)}\in\mathcal{H}^{(1)}} {\sum_{\textbf{t}^{(1)}\in\mathcal{T}^{(1)}} \textbf{D}}.
 	\end{aligned}
 	%.\nonumber
 	\end{equation*}
 	
 	\begin{figure}[!h]
 		\centering
 		\includegraphics[height=11cm, width=16cm,keepaspectratio]{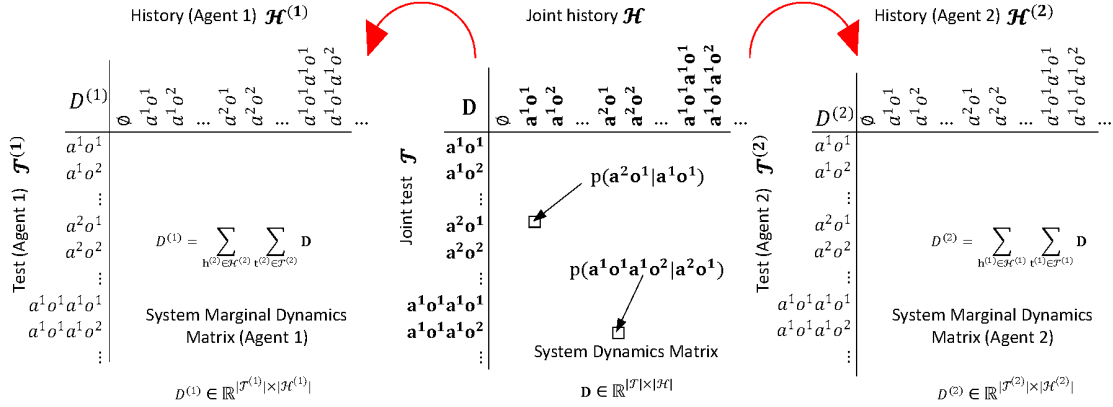}
 		\caption{Diagram of system marginal dynamics matrices obtained by the system dynamics matrix of 2-agent system}
 		\label{Fig.smdij}
 	\end{figure}
 	
 	Correspondingly, the system PSR model in the perspective of two agents has the following parameters.
 	\begin{equation}
 	\left\lbrace
 	\begin{aligned}
 	<\mathcal{A}^{(1)}, \mathcal{O}^{(1)}, \textbf{Q}^{(1)}, \{\sum_{\textbf{ao}^{(2)}} \bm_{\textbf{ao}}\},\{\sum_{\textbf{ao}^{(2)}} M_{\textbf{ao}}\}, \sum_{\textbf{Q}^{(2)}} p(\textbf{Q}|\phi)>\\
 	<\mathcal{A}^{(2)}, \mathcal{O}^{(2)}, \textbf{Q}^{(2)}, \{\sum_{\textbf{ao}^{(1)}} \bm_{\textbf{ao}}\},\{\sum_{\textbf{ao}^{(1)}} M_{\textbf{ao}}\}, \sum_{\textbf{Q}^{(1)}} p(\textbf{Q}|\phi)>
 	\end{aligned}
 	\right..\nonumber
 	\end{equation}
 	%\begin{flalign}
 	%<\mathcal{A}^{(1)}, \mathcal{O}^{(1)}, \textbf{Q}^{(1)}, \{\sum_{\textbf{ao}^{(2)}} \bm_{\textbf{ao}}\},\{\sum_{\textbf{ao}^{(2)}} M_{\textbf{ao}}\}, \sum_{\textbf{Q}^{(2)}} p(\textbf{Q}|\phi)>\\
 	%<\mathcal{A}^{(2)}, \mathcal{O}^{(2)}, \textbf{Q}^{(2)}, \{\sum_{\textbf{ao}^{(1)}} \bm_{\textbf{ao}}\},\{\sum_{\textbf{ao}^{(1)}} M_{\textbf{ao}}\}, \sum_{\textbf{Q}^{(1)}} p(\textbf{Q}|\phi)>
 	%\end{flalign}
 	
 	Similarly, for a general multi-agent system, the system marginal dynamics matrices can be obtained by the system dynamics tensor $\mathcal{D}$
 	and the system dynamics matrix $\textbf{D}$ respectively, and the PSR model of each agent can also be directly obtained from the learned multi-agent PSR model, namely:
 	\begin{equation}
 	\left\lbrace
 	\begin{aligned}
 	D^{(1)}&=\sum_{\textbf{h}^{(\neq 1)}\in\mathcal{H}^{(\neq 1)}} {\sum_{\textbf{t}^{(\neq 1)}\in\mathcal{T}^{(\neq 1)}} \mathcal{D}}\\
 	&\vdots\\
 	D^{(N)}&=\sum_{\textbf{h}^{(\neq N)}\in\mathcal{H}^{(\neq N)}} {\sum_{\textbf{t}^{(\neq N)}\in\mathcal{T}^{(\neq N)}}\mathcal{D}}
 	\end{aligned}
 	\right.
 	~\text{and}\quad
 	\left\lbrace
 	\begin{aligned}
 	D^{(1)}&=\sum_{\textbf{h}^{(\neq 1)}\in\mathcal{H}^{(\neq 1)}} {\sum_{\textbf{t}^{(\neq 1)}\in\mathcal{T}^{(\neq 1)}} \textbf{D}}\\
 	&\vdots\\
 	D^{(N)}&=\sum_{\textbf{h}^{(\neq N)}\in\mathcal{H}^{(\neq N)}} {\sum_{\textbf{t}^{(\neq N)}\in\mathcal{T}^{(\neq N)}}\textbf{D}}
 	\end{aligned}
 	\right.,\nonumber
 	\end{equation}
 	and
 	\begin{equation}
 	\left\lbrace
 	\begin{aligned}
 	<\mathcal{A}^{(1)}, \mathcal{O}^{(1)}, \textbf{Q}^{(1)}, \{\sum_{\textbf{ao}^{(\neq 1)}} \bm_{\textbf{ao}}\},\{\sum_{\textbf{ao}^{(\neq 1)}} M_{\textbf{ao}}\}, \sum_{\textbf{Q}^{(\neq 1)}} p(\textbf{Q}|\phi)>\\
 	\vdots\hspace{20em} \\
 	<\mathcal{A}^{(N)}, \mathcal{O}^{(N)}, \textbf{Q}^{(N)}, \{\sum_{\textbf{ao}^{(\neq N)}} \bm_{\textbf{ao}}\},\{\sum_{\textbf{ao}^{(\neq N)}} M_{\textbf{ao}}\}, \sum_{\textbf{Q}^{(\neq N)}}p(\textbf{Q}|\phi)>\\
 	\end{aligned}
 	\right..\nonumber
 	\end{equation}
 	Accordingly we can use these matrices and the PSR model for individual agent planning.}
 
 \section{Learning 2-agent PSR via Tensor Decomposition}\label{sec:modeling}
 In this section, we propose a new framework for learning 2-agent PSR via tensor decomposition.
 We elaborate how to obtain the prediction parameters $\tilde{m}_{ij}$ and the state vector $x_k$ of the PSR model through CP decomposition (CP) and Tucker decomposition (TD) respectively, and then use a linear regression to learn the transition parameters $\tilde{M}_{\textbf{ao}}$ of the model from the training data.
 %To begin with, analogous to the system dynamics matrix for single agent,
 %we construct the dynamics tensor $\mathcal{D}\in \mathbb{\R}^{n_1\times n_2 \times n_3}$ for the
 %two agents, whose element is $\mathcal{D}_{ijk} = p(t_i,t_j|\textbf{h}_k)$, where $t_i$ and $t_j$ are the tests for each agent and $\textbf{h}_k$ is their joint history.
 %, that is, to complete the whole learning progress of PSRs model.
 
 \subsection{Learning Prediction Parameters and State Vectors}\label{keymij}
 
 {After obtaining interactive data $\Phi_d$ between two agents, we construct the system dynamics tensor $\mathcal{D}\in \mathbb{\R}^{n_1\times n_2 \times n_3}$, whose element is $\mathcal{D}_{ijk} = p(t_i^{(1)}t_j^{(2)}|\textbf{h}_k)$, where $t_i^{(1)}$ is the $i$-th test sequence of the $1$st agent's test set $\mathcal{T}^{(1)}$, $t_j^{(2)}$ is the $j$-th test sequence of the $2$nd agent's test set $\mathcal{T}^{(2)}$ and $\textbf{h}_k\in \mathcal{H}$($k\in[1,|\mathcal{H}|)$) is the given joint history at time step $s=|\textbf{h}_k|$.}
 We use two main tensor decomposition approaches on tensor $\mathcal{D}$ for learning the prediction parameters and the state vector respectively.
 %, see Steps 1-2 of Fig. \ref{Fig.CPTUCKER}.
 %Most notations according to tensor operations follow from \cite{Kolda2009Tensor}.
 
 \subsubsection{CP Decomposition Learning Method}\label{keycp}
 %According to the definition of CP decomposition, that is:
 The CP decomposition decomposes a tensor into a sum of rank-one
 tensors that can be concisely written in Eq.~(\ref{eq:61}).
 
 \begin{equation} \label{eq:61}
 \mathcal{D}\approx [\textbf{\(\lambda\)};A,B,C]\equiv \sum^{R}_{r=1} \lambda_r a_r \circ b_r \circ c_r,
 \end{equation}
 where $R$ is a positive integer, ``$\circ$'' denotes outer product of vectors, and $\lambda_r \in \R$,
 %$[\textbf{\(\lambda\)};\textbf{A},\textbf{B},\textbf{C}]$ is a simplified notation of tensor's CP decomposition\cite{Kolda2009Tensor},
 $a_r \in \R^{n_1}$, $b_r \in \R^{n_2}$ and $c_r \in \R^{n_3}$ for $ r = 1,2, \ldots , R$.
 The factor matrices $A$, $B$ and $C$ {consist of the vectors}, i.e.,
 $A=[a_1\ a_2\  \cdots \ a_R]\in \mathbb{\R}^{n_1 \times R}$, $B=[b_1\  b_2\  \cdots\  b_R]\in \mathbb{\R}^{n_2 \times R}$ and $C=[c_1\  c_2\  \cdots\  c_R]\in \mathbb{\R}^{n_3 \times R}$.
 For any $ i\in\{1,\ldots,n_1\}$, $j\in\{1,\ldots,n_2\}$ and $k\in\{1,\ldots,n_3\}$, we observe that each element of the tensor $\mathcal{D}$ can be written in Eq.~(\ref{eq:62}).
 
 \begin{equation}
 \mathcal{D}_{ijk} =  \sum^{R}_{r=1} \lambda_r A_{i r}  B_{j r}  C_{k r},
 \label{eq:62}
 \end{equation}
 %where, the vector  $a_{i*}$ is the $i$-th row vector of matrix $A$, the vector  $b_{j*}$ is the $j$-th row vector of  matrix $B$ and  the vector  $c_{k*}$ is the $k$-th row vector of  matrix $C$.
 where $A_{i r}$ is the $(i,r)$-th element of matrix $A$, and likewise for $B_{j r}$ and $C_{k r}$, for $r=1,2,\ldots, R$.
 
 Let $x_k =[x_k(1) \cdots x_k(R)]\in \mathbb{\R}^{1 \times R}$, $k\in\{1,\ldots,n_3\}$ be the $k$-th row vector $C_{k:}$.
 {Since the joint histories of dynamic system stored in the 3rd dimension of the tensor $\mathcal{D}$ are compressed in the matrix $C$, its row vector $x_k$ is a summary of joint history $\textbf{h}_k\in \mathcal{H}$($k\in[1,|\mathcal{H}|$) and can be considered as a compressed version of the system state vector $p(\textbf{Q} |\textbf{h}_k)$ at time step $s=|\textbf{h}_k|$. On the other hand,}
 for any $ r\in\{1,2,\ldots,R\}$, we define the scalar $\tilde{m}_{ij}(r)=\lambda_r A_{ir}  B_{jr}$, and then construct the column vector $\tilde{m}_{ij}={[ \tilde{m}_{ij}(1) \  \tilde{m}_{ij}(2)\ \cdots\  \tilde{m}_{ij}(R) ]}^T \in \mathbb{\R}^{R}$. Thus, we have
 %\begin{flalign}\label{eq:640}
 %m_{ij}(r)= \lambda_r a_{ir}  b_{jr},\nonumber\hspace{2em}\\
 %m_{ij}=(\lambda *A_{i:} * B_{j:})^T,
 %\end{flalign}
 \begin{equation}\label{eq:640}
 \tilde{m}_{ij}=(\lambda \ast A_{i:} \ast B_{j:})^T,
 \end{equation}
 where $\lambda =[\lambda_1 \cdots \lambda_R]\in \mathbb{\R}^{1 \times R}$, ``$\ast$'' denotes the Hadamard product~(vector element-wise product),
 $A_{i:}$ denotes the $i$-th row vector of $A$ and likewise for $B_{j:}$.
 Then, from (\ref{predictivetpsr}), we rewrite Eq.~(\ref{eq:62}) as
 %\begin{equation}
 %\mathcal{D}_{i,j,k} =  \sum^{R}_{r=1} m_{ij}(r)  x_k(r) = x_k m_{ij}
 %\label{eq:64}
 %\end{equation}
 \begin{equation}%\label{eq:64}
 \mathcal{D}_{ijk} = \sum^{R}_{r=1} \tilde{m}_{ij}(r)  x_k(r) =  x_{k}(\lambda \ast A_{i:} \ast B_{j:})^T = x_k \tilde{m}_{ij}.
 \nonumber
 \end{equation}
 Hence, the prediction parameters $\tilde{m}_{ij}$ and the compressed state vector $x_k$ are obtained from the tensor decomposition results. Remark that we do not compute $x_k$ directly from Eq.~(\ref{eq:631}), and $p(\textbf{Q}|\textbf{h}_k)$ and $F$ are unknown currently.
 %According to the Eq.~(\ref{eq:631}) in the theory of TPSR, the $x_k$ can be obtained by the system prediction vector $p(\textbf{Q} |\textbf{h}_k)$ multiply a projection matrix $F$, i.e., $x_k = p(\textbf{Q}|\textbf{h}_k)F$, where the $p(\textbf{Q}|\textbf{h}_k)$ and $F$ are unknown, and it not necessary to try to obtain them here.
 
\begin{figure}[!h]
 	\centering
 	\includegraphics[height=12cm, width=15cm,keepaspectratio]{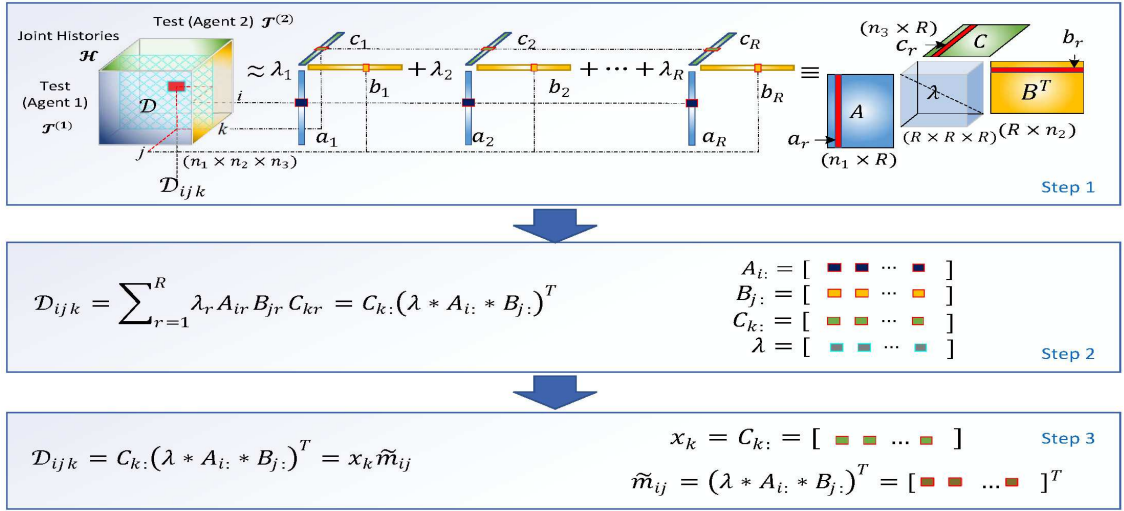}
 	\caption{The process of learning the prediction parameters $\tilde{m}_{ij} $  and the compressed state vector $x_k$  by CP decomposition}
 	\label{Fig.cpdijk}
 \end{figure}

 The computation process of learning the prediction
 parameters $\tilde{m}_{ij}$ and the compressed state vector $x_k$ by CP decomposition is shown in Fig. \ref{Fig.cpdijk}.
 After applying the CP decomposition to the system dynamics tensor, we have 3 factor matrices and a diagonal tensor in Step 1. For each element $\mathcal{D}_{ijk}$ in tensor $\mathcal{D}$, it can be realized by the CP decomposition results shown in Step 2. Hence, we can deduce the prediction parameters from $\mathcal{D}_{ijk}$ in Step 3.
 Moreover, the system state vector is also obtained for a further use.
 \subsubsection{Tucker Decomposition Learning Method}
 
 The Tucker decomposition decomposes a tensor
 into a core tensor multiplied (or transformed) by a matrix along each mode, i.e.,
 \begin{flalign}
 \mathcal{D} \approx [\mathcal{G};A,B,C]\equiv  \mathcal{G}\times_1 A \times_2 B \times_3 C\hspace{3.5em} \nonumber\\
 = \sum^{P}_{p=1} \sum^{Q}_{q=1}\sum^{R}_{r=1} g_{pqr}a_p \circ b_q \circ c_r,
 \label{eq:71}
 \end{flalign}
 where ``$\times_k$'' denotes $k$-mode
 product of tensor $\mathcal{G}$ by {a matrix with appropriate dimensions} and $P, Q, R$ are all positive integers.
 %$[\mathcal{G};\textbf{A},\textbf{B},\textbf{C}]$ is a simplified notation of tensor's Tucker decomposition\cite{Kolda2009Tensor}.
 Usually $P<n_1$, $Q<n_2$ and $R<n_3$, the core tensor $\mathcal{G}$ can be thought of as a compressed version of $\mathcal{D}$.
 %$\forall r\in \{1,2, \dots ,R\}$, $a_r \in \R^{n_1}$,$b_r \in \R^{n_2}$ and $c_r \in \R^{n_3}$.
 The factor matrices $A, B$ and $C$ can be computed in Eq. \ref{eq:72}.
 $A=[a_1\ a_2\  \cdots\ a_P]\in \mathbb{\R}^{n_1 \times P}$,
 $B=[b_1\  b_2\  \cdots\  b_Q]\in \mathbb{\R}^{n_2 \times Q}$ and
 $C=[c_1\  c_2\  \cdots\  c_R]\in \mathbb{\R}^{n_3 \times R}$.
 Similarly, each element of the tensor $\mathcal{D}$ can be written as
 \begin{equation}
 \mathcal{D}_{ijk} =  \sum^{P}_{p=1} \sum^{Q}_{q=1}\sum^{R}_{r=1} g_{pqr} A_{ip}  B_{jq}  C_{kr}.
 \label{eq:72}
 \end{equation}
 %Since the tensor of the dynamic system $\mathcal {D}$is compressed in $C $\cite{Zheng2016Topic}, it can be considered that the row vector $x_k$is the system prediction vector $P (Q | H_k) $ which is transformed by the projection matrix $\textbf {T}$, and is the characteristic representation of the system state information.
 %\begin{equation}
 %x_k = P(Q|H_k)\textbf{T},
 %\label{eq:73}
 %\end{equation}
 %where the vector $a_{i*}$ is the $i$-th row vector of matrix $A$, and likewise for the vectors $b_{j*}$ and $c_{k*}$.

 %\textcolor{green}{Since the joint histories of dynamic system stored in the 3rd dimension of the dynamic system's tensor $\mathcal{D}$ is compressed in $C$, whose row vector $x_k$ can be considered as a compressed version of the system state vector at the time step $k$. According to the Eq.~(\ref{eq:631}) in the theory of TPSR, the $x_k$ can be obtained by the system prediction vector $p(\textbf{Q} |\textbf{h}_k)$ multiply a projection matrix $F$, i.e., $x_k = p(\textbf{Q}|\textbf{h}_k)F$, where the $p(\textbf{Q}|\textbf{h}_k)$ and $F$ are unknown, and it not necessary to try to obtain them here.}
 {Let $x_k =[x_k(1) \cdots x_k(R)]\in \mathbb{\R}^{1 \times R}$, $k\in\{1,\ldots,n_3\}$ be the $k$-th row vector $C_{k:}$.
 	As discussed in Section \ref{keycp}, the row vector of $C$ is the compressed state vector $x_k$ of the TPSR model.}
 For any $ r\in\{1,2,\ldots,R\}$, we define the scalar $\tilde{m}_{ij}(r)=\sum^{P}_{p=1} \sum^{Q}_{q=1} g_{pqr} A_{ip}  B_{jq}$, and construct the column vector $\tilde{m}_{ij}={[ \tilde{m}_{ij}(1) \  \tilde{m}_{ij}(2)\ \cdots\  \tilde{m}_{ij}(R) ]}^T \in \mathbb{\R}^{R}$. Thus, we get
 \begin{flalign}
 %m_{ij}(r) =  \sum^{P}_{p=1} \sum^{Q}_{q=1} g_{pqr} a_{ip} b_{jq}\nonumber,\\
 \tilde{m}_{ij} =\mathcal{G}_{(3)}(B_{j:}\otimes A_{i:})^T,%\hspace{0.5em}
 \label{eq:740}
 \end{flalign}
 where ``$\otimes$'' denotes Kronecker product, and $\mathcal{G}_{(3)}$ is the mode-3 unfolding of the core tensor $\mathcal{G}$.
 Hence, from Eq.~(\ref{predictivetpsr}), Eq.~(\ref{eq:72}) becomes
 \begin{equation}
 \mathcal{D}_{ijk} =  \sum^{R}_{r=1} \tilde{m}_{ij}(r)  x_k(r) = x_k \tilde{m}_{ij}.
 %\label{eq:74v}
 \nonumber
 \end{equation}
 
\begin{figure}[!h]
 	\centering
 	\includegraphics[height=12cm, width=15cm,keepaspectratio]{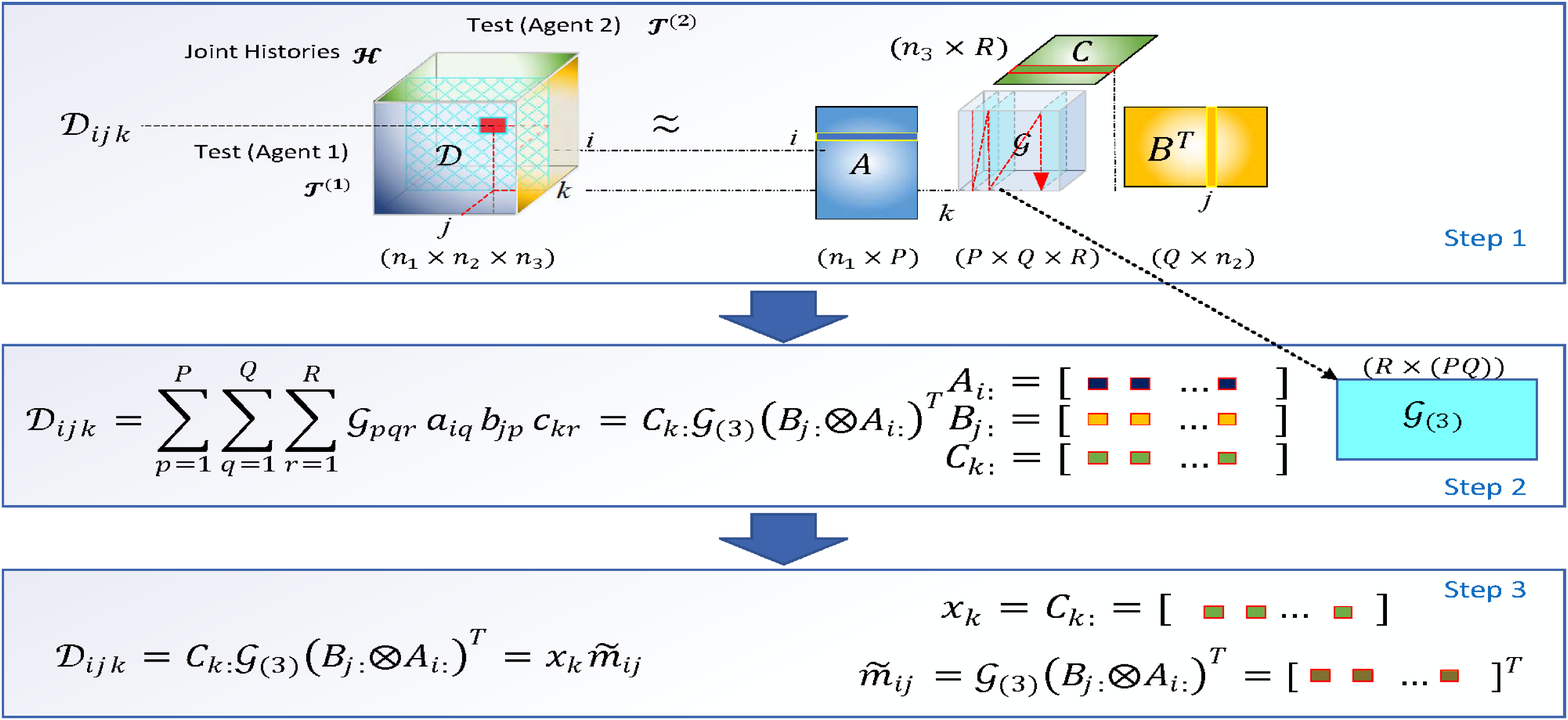}
 	\caption{The process of learning the prediction parameters $\tilde{m}_{ij}$ and the compressed state vector $x_k$ by Tucker decomposition}
 	\label{Fig.tddijk}
 \end{figure}

 %{
 %There are two equivalent ways for solving $m_{ij}$, which is depicted in Fig. \ref{Fig.tuckermijr}. In method one, we can follow the Tucker decomposition theory for computing the  element of $m_{ij}$, namely $m_{ij}(r)$, which is equal to $r$-th lateral slice matrix $\mathcal{G}_{::,r}$ of core tensor $\mathcal{G}$ multiply with the $i$-th and $j$-th row in factor matrixes $A$ and $B$  in the left side and right side.
 %In the alternative one, we can unfold the core tensor $\mathcal{G}$ in 3rd-dimension for Mode-$3$ unfolding matrix $\mathcal{G}_{(3)}$ and apply Kronecker product to the $i$-th and $j$-th row in factor matrixes $A$ and $B$ for a expanding vector. Then, we can get the vector $m_{ij}$ through multiplying the matrix $\mathcal{G}_{(3)}$ with the expanding vector.
 %}
 {
 	The computation  process of learning the prediction
 	parameters $\tilde{m}_{ij}$ and the compressed state vector $x_k$ by Tucker decomposition is shown in Fig. \ref{Fig.tddijk}.
 	Similarly, we obtain 3 factor matrices and a core tensor after applying the Tucker decomposition in Step 1.
 	Then in Step 2, each element $\mathcal{D}_{ijk}$ can be reorganized by the Tucker decomposition results. Hence, we can deduce the prediction parameters in Step 3 and the system state vector as well.
 }
 %\begin{figure}[!h]
 %	\centering
 %	\includegraphics[height=12cm, width=15cm,keepaspectratio]{5/TUCKERmijrNN.eps}
 %	\caption{The equivalent process of solving $m_{ij}$}
 %	\label{Fig.tuckermijr}
 %\end{figure}
 
 We may add some proper constraints to Equations~(\ref{eq:61}) and~(\ref{eq:71}) to ensure
 solution uniqueness or algorithm convergence in the tensor decompositions.
 We assume that the norm of all columns of $A, B$ and $C$ are 1 for Eq.~(\ref{eq:61}),
 and $A, B$ and $C$ are column-wise orthonormal for problem (\ref{eq:71}).
 If we further add non-negative constraints on $A, B$ and $C$, the problems become
 non-negative CP decomposition\,(NCP) and non-negative Tucker decomposition\,(NTD), respectively.
 There are many methods devoted for solving these NP-hard problems (in general), and the corresponding
 algorithms may converge to a stationary point and enjoy convergence guarantee under certain conditions, see e.g.,
 \cite{Kolda2009Tensor}, \cite{Xu2015A} and \cite{Kim2014Algorithms}.
 
 %{Biyang: we can delete this paragraph if you have stated it in the previous paragraphs. Hence, it depends on you how to organize them.} From the decomposition results, we can derive the prediction parameters
 %$\tilde{m}_{ij}$ for the PSR model, i.e., Eqs.~(\ref{eq:640}) and (\ref{eq:740}).
 %Since the histories of the dynamic system $\mathcal {D}$ are compressed in factor matrix $C$, its row vector (i.e., $x_k$) is considered as the compressed version of system prediction vector $P (\textbf{Q}|\textbf{h}_k) $, which is the characteristic representation of the system state information.

 \subsection{Learning Transition Parameters}
\begin{figure}[t]
 	\centering
 	\includegraphics[height=15cm, width=15cm,keepaspectratio]{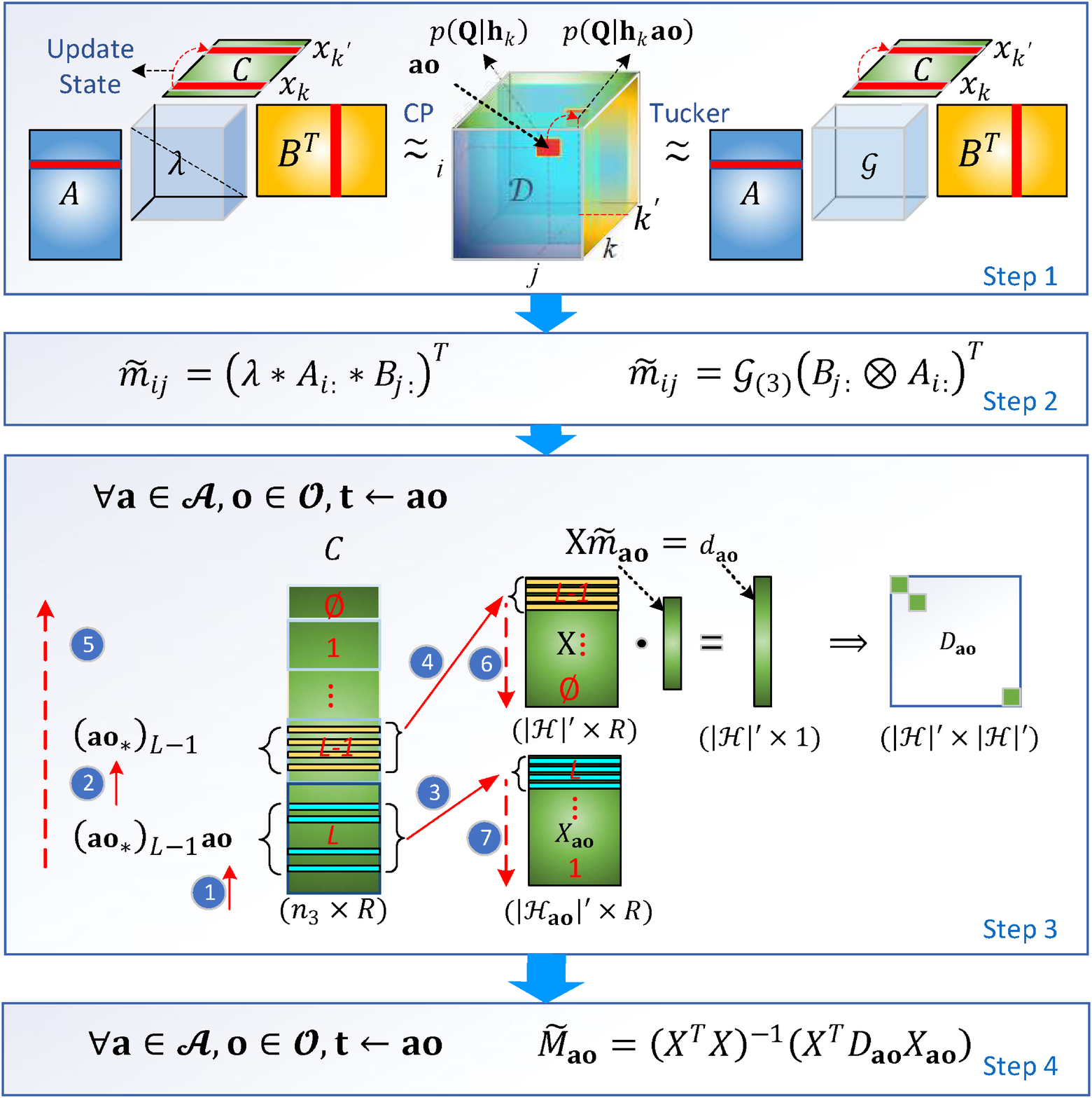}
 	\caption{Tensor decomposition for learning 2-agent PSR}
 	\label{Fig.CPTUCKER}
 \end{figure}

 {In Section \ref{keymij}, we have derived the prediction parameters $\{\tilde{m}_{ij}\}$ for the TPSR model through two tensor decomposition methods respectively, see Steps 1-2 of Fig. \ref{Fig.CPTUCKER}. From the analysis in Section \ref{sec:preparations}, we can find the subset of one-step projection vectors  $\{\tilde{m}_{\textbf{ao}}\}$ from  the prediction parameter set $\{\tilde{m}_{ij}\}$ for each $\textbf{a}\textbf{o} \in \mathcal{A}\times \mathcal{O}$, which is important in learning the model transition parameters $M_{\textbf{ao}}$.
 	In addition, we know that it is very difficult and not necessary to get the transformation matrix ${F}$. Therefore, we try to learn the model transition parameters $M_{\textbf{ao}}$ through a linear regression from the training data, see Steps 3-4 of Fig. \ref{Fig.CPTUCKER}.}
 
 {
 	For any $\textbf{a} \in \mathcal{A},\textbf{o} \in \mathcal{O}$, we find all the joint histories $\textbf{h} \in \mathcal{H}$ ending with action-observation $\textbf{ao}$ and construct a subset of joint history set $\mathcal{H}$, i.e., $\mathcal{H}^{\prime}_{\textbf{ao}} = \{\textbf{h}|\textbf{h} = \textbf{h}^{\prime} \textbf{ao},\textbf{h} \in \mathcal{H},\textbf{h}^{\prime} \in \mathcal{H}\}$. Moreover, for every joint history $\textbf{h}^{\prime}\in \mathcal{H}^{\prime}_{\textbf{ao}}$, we cut off the action-observation $\textbf{ao}$, obtain the new joint history $\textbf{h}=\textbf{h}^{\prime}\setminus \textbf{ao}$ which absolutely belongs to the joint history set $\mathcal{H}$, and then construct a subset of joint history set $\mathcal{H}$, i.e., $\mathcal{H}^{\prime} = \{\textbf{h}|\textbf{h} = \textbf{h}^{\prime}\setminus \textbf{ao}, \textbf{h} \in \mathcal{H},\textbf{h}^{\prime} \in \mathcal{H}^{\prime}_{\textbf{ao}}\}$. We see that the size of the joint history set $\mathcal{H}^{\prime}$ is equal to that of the joint history set $\mathcal{H}^{\prime}_{\textbf{ao}}$, i.e., $|\mathcal{H}^{\prime}|=|\mathcal{H}^{\prime}_{\textbf{ao}}|$.
 	%And for any joint history $\textbf{h}_{k}\in\mathcal{H}^{\prime}$ and $\textbf{h}_k^{\prime} \in\mathcal{H}^{\prime}_{\textbf{ao}}$, the equation $\textbf{h}_k^{\prime} = \textbf{h}_{k}\textbf{ao}$ is hold.
 	
 	Let $\mathcal{I}(\mathcal{H}^{\prime},\mathcal{H})$ be an index set that records the indices of each element of the subset $\mathcal{H}^\prime$ in the set $\mathcal{H}$. Particularly,
 	$\mathcal{I}(\textbf{h}_k\in\mathcal{H}^{\prime},\mathcal{H})$ is an index that records the index of joint history $\textbf{h}_k\in\mathcal{H}^{\prime}$ in the set $\mathcal{H}$.
 	Given $x_{k}~(k\in\mathcal{I}(\mathcal{H}^{\prime},\mathcal{H}))$, we multiply both sides of Eq.~(\ref{eq:661}) by
 	${x_{k}  \tilde{m}_{\textbf{ao}}}$, and obtain a series of equations with a size equal to $|\mathcal{H}^\prime|$.
 	%and arrange the equations according to the length of joint history $\textbf{h}_k\in\mathcal{H}^{\prime}$,.
 	Thus, we have
 	\begin{equation}\label{eq:677}
 	\left\lbrace
 	\begin{aligned}
 	x_{\mathcal{I}(\textbf{h}_1^{\prime}\in\mathcal{H}^{\prime}_{\textbf{ao}},\mathcal{H})}  x_{\mathcal{I}(\textbf{h}_1\in\mathcal{H}^{\prime},\mathcal{H})} \tilde{m}_{\textbf{ao}}&=x_{\mathcal{I}(\textbf{h}_1\in\mathcal{H}^{\prime},\mathcal{H})}\tilde{M}_{\textbf{ao}}\\
 	x_{\mathcal{I}(\textbf{h}_2^{\prime}\in\mathcal{H}^{\prime}_{\textbf{ao}},\mathcal{H})} x_{\mathcal{I}(\textbf{h}_2\in\mathcal{H}^{\prime},\mathcal{H})} \tilde{m}_{\textbf{ao}}&=x_{\mathcal{I}(\textbf{h}_2\in\mathcal{H}^{\prime},\mathcal{H})}\tilde{M}_{\textbf{ao}}\\
 	\vdots\hspace{2em} &= \hspace{1em} \vdots\\
 	x_{\mathcal{I}(\textbf{h}_{|\mathcal{H}^{\prime}_{\textbf{ao}}|}^{\prime}\in\mathcal{H}^{\prime}_{\textbf{ao}},\mathcal{H})} x_{\mathcal{I}(\textbf{h}_{|\mathcal{H}^{\prime}|}\in\mathcal{H}^{\prime},\mathcal{H})}  \tilde{m}_{\textbf{ao}}&=x_{\mathcal{I}(\textbf{h}_{|\mathcal{H}^{\prime}|}\in\mathcal{H}^{\prime},\mathcal{H})} \tilde{M}_{\textbf{ao}}
 	\end{aligned}
 	\right.,\nonumber
 	\end{equation}
 	where $\textbf{h}_k^{\prime} = \textbf{h}_k \textbf{ao} $, $k=1,2,\ldots, |\mathcal{H}^{\prime}|  $ and the subscripts $\mathcal{I}(\textbf{h}_k^{\prime}\in\mathcal{H}^{\prime}_{\textbf{ao}},\mathcal{H})$ and $\mathcal{I}(\textbf{h}_k\in\mathcal{H}^{\prime},\mathcal{H})$ indicate the row indices of the factor matrix $C$.
 	%, i.e. $\mathcal{I}(\textbf{h}_k\in\mathcal{H}^{\prime},\mathcal{H})$ is a indicator returns the index of joint history $\textbf{h}_k\in\mathcal{H}^{\prime}$ in joint history set $\mathcal{H}$ which can map the index of its element to the row index of matrix $C$.
 	Then, we extract the row $x_{\mathcal{I}(\textbf{h}_{k}\in\mathcal{H}^{\prime},\mathcal{H})}$ from $C$ according to the index $\mathcal{I}(\textbf{h}_{k}\in\mathcal{H}^{\prime},\mathcal{H})$, and construct the system state matrix $X\in \mathbb{\R}^{|\mathcal{H}^{\prime}|\times R}$, i.e., $X\leftarrow C_{\mathcal{I}(\mathcal{H}^{\prime},\mathcal{H})}$.
 	Similarly, we extract the row $x_{\mathcal{I}(\textbf{h}_k^{\prime} \in\mathcal{H}^{\prime}_{\textbf{ao}},\mathcal{H})}$ from $C$
 	and construct the one-step extension system state matrix $X_{\textbf{ao}}\in \mathbb{\R}^{|\mathcal{H}^{\prime}_{\textbf{ao}}|\times R}$,
 	i.e., $X_{\textbf{ao}}\leftarrow C_{\mathcal{I}(\mathcal{H}_{\textbf{ao}}^{\prime},\mathcal{H})}$.
 	For simplicity, we denote $C_{\mathcal{I}(\mathcal{H}^{\prime},\mathcal{H})}$  and $C_{\mathcal{I}(\mathcal{H}^{\prime}_{\textbf{ao}},\mathcal{H})}$ as  $C_{\mathcal{H}^{\prime}}$ and $C_{\mathcal{H}^{\prime}_{\textbf{ao}}}$, respectively.
 	Hence, we have}
 %from the factor matrix $C$,  and $X_{\textbf{ao}}\leftarrow C_{\mathcal{I}(\mathcal{H}^{\prime}_{\textbf{ao}},\mathcal{H})}$,
 %where $C_{\mathcal{I}(\mathcal{H}^{\prime},\mathcal{H})}$ extracts each row $x_{\mathcal{I}(\textbf{h}_{k}\in\mathcal{H}^{\prime},\mathcal{H})} \in C$
 %with subscript ${\mathcal{I}(h_{k}\in\mathcal{H}^{\prime},\mathcal{H})}$ which is the joint history $\textbf{h}_k \in \mathcal{H}^{\prime}$'s index in joint history set $\mathcal{H}$,
 %For the set of rows in $C$ can be explicitly indicated by $\mathcal{H}$, $\mathcal{H}^{\prime}$ and $\mathcal{H}^{\prime}_{\textbf{ao}}$, it means we can  drop $C_{\mathcal{I}(\mathcal{H}^{\prime},\mathcal{H})}$ to $C_{\mathcal{H}^{\prime}}$ and $C_{\mathcal{I}(\mathcal{H}^{\prime}_{\textbf{ao}},\mathcal{H})}$ to $C_{\mathcal{H}^{\prime}_{\textbf{ao}}}$ for simplicity.
 %Thus, we have $X\leftarrow C_{\mathcal{H}^{\prime}}$ and $X_{\textbf{ao}}\leftarrow C_{\mathcal{H}^{\prime}_{\textbf{ao}}}$, namely
 \begin{flalign}%\label{eq:X}
 X =
 \begin{bmatrix}
 x_{\mathcal{I}(\textbf{h}_1\in\mathcal{H}^{\prime},\mathcal{H})} \\
 x_{\mathcal{I}(\textbf{h}_2\in\mathcal{H}^{\prime},\mathcal{H})} \\
 \vdots \\
 x_{\mathcal{I}(\textbf{h}_{|\mathcal{H}^{\prime}|}\in\mathcal{H}^{\prime},\mathcal{H})}
 \end{bmatrix}
 , ~\text{and} \quad
 X_{\textbf{ao}} =
 \begin{bmatrix}
 x_{\mathcal{I}(\textbf{h}_1^{\prime}\in\mathcal{H}^{\prime}_{\textbf{ao}},\mathcal{H})} \\
 x_{\mathcal{I}(\textbf{h}_2^{\prime}\in\mathcal{H}^{\prime}_{\textbf{ao}},\mathcal{H})} \\
 \vdots \\
 x_{\mathcal{I}(\textbf{h}_{|\mathcal{H}^{\prime}_{\textbf{ao}}|}^{\prime}\in\mathcal{H}^{\prime}_{\textbf{ao}},\mathcal{H})}
 \end{bmatrix}.\nonumber
 \end{flalign}
 Let $d_{\textbf{ao}}=X \tilde{m}_{\textbf{ao}}$, and $D_{\textbf{ao}} $ be the matrix with $d_{\textbf{ao}}$ on its diagonal. Thus, Eq.~(\ref{eq:677}) can be written as
 \begin{flalign}\label{eq:67}
 D_{\textbf{ao}} X_{\textbf{ao}}= X \tilde{M}_{\textbf{ao}}.
 \end{flalign}
 The value $\tilde{M}_{\textbf{ao}}$ will be exactly true if we have a perfect estimation of $X$, $X_{\textbf{ao}}$ and $\tilde{m}_{\textbf{ao}}$ from infinite training data.
 With finite training data, it is generally not possible to have precise $\tilde{M}_{\textbf{ao}}$ in Eq.~(\ref{eq:67}). Hence we
 %pre-multiply both sides of Eq.~(\ref{eq:67}) by $D_{\textbf{ao}}$, and
 formulate the following optimization problem below
 \begin{flalign}
 \min\limits_{\tilde{M}}  \dfrac{1}{2} \|D_{\textbf{ao}} X_{\textbf{ao}}- X \tilde{M} \|_F^2,\nonumber
 \end{flalign}
 where $\|\cdot\|_F$ is the Frobenius norm of a matrix.
 Taking derivative on $\tilde{M}$, we have the optimal solution
 \begin{flalign}\label{eq:691}
 \tilde{M}^*_{\textbf{ao}} =  {(X^T X)}^{-1}(X^T D_{\textbf{ao}} X_{\textbf{ao}}) .
 \end{flalign}
 
 Fig. \ref{Fig.CPTUCKER}  shows the detailed calculations of $\tilde{m}_{\textbf{ao}}$, $\tilde{M}_{\textbf{ao}}$, and $X$, $X_{\textbf{ao}}$ and $D_{\textbf{ao}}$ as well. As shown in the figure, we offer two frameworks for decomposing the tensor, and obtain a state matrix $C$, which means that the state vector of state matrix $C$ is updated while the system updates along the red dashed link after receiving the joint action-observation \textbf{ao}. After applying tensor decomposition~(either CP or Tucker) to the system dynamics tensor $\mathcal{D}$~(Step 1), we immediately obtain the prediction parameters $\tilde{m}_{\textbf{ao}}$ in Step 2. In Step 3, starting from the joint histories of the longest action-observation sequences~(L),
 the process goes: {\textcircled{\small{1}}}
 finds out all the rows of $C$ corresponding to the action-observation sequence ending with $\textbf{ao}${, which can be indexed by the elements of joint history set $\mathcal{H}^{\prime}_{\textbf{ao}}$;} {\textcircled{\small{2}}} finds out the corresponding rows of $C$ after deleting $\textbf{ao}${, which can be indexed by the elements of joint history set $\mathcal{H}^{\prime}$;} {\textcircled{\small{3}}} puts the row vectors from {\textcircled{\small{1}}} into the matrix $X_{\textbf{ao}}$; {\textcircled{\small{4}}} puts the row vectors from {\textcircled{\small{2}}} into the matrix $X$; {\textcircled{\small{5}}}-{\textcircled{\small{7}}} repeat from {\textcircled{\small{1}}} to {\textcircled{\small{4}}} until the empty sequences appear. Finally, we calculate $\tilde{M}_{\textbf{ao}}$ in Eq.~(\ref{eq:691}) in Step 4.
 
 \section{Multi-agent PSR via Tensor Decomposition}\label{sec:modeling2}
 We extend the learning 2-agent PSR model to the case of multiple agents.
 Given a dynamic system has $N$ agents, we have  tensor $\mathcal{D}\in \mathbb{\R}^{n_1\times n_2 \times n_3 \times \cdots  \times  n_{N+1}}$, whose element is $\mathcal{D}_{i_1 i_2\dots i_N k} =p(t_{i_1}^{(1)}\cdots t_{i_N}^{(N)}|\textbf{h}_{k})$
 %, where $t_{i_1}^{(1)}$, $t_{i_2}^{(2)}$, $\ldots$, $t_{i_N}^{(N)}$ are the tests for each agent and $\textbf{h}_k\in \mathcal{H}$($k\in[1,|\mathcal{H}|$) is the given joint history at time step $s=|\textbf{h}_k|$
 . Its CP decomposition:
 \begin{flalign}
 \mathcal{D} &\approx [\textbf{\(\lambda\)};A^{(1)},A^{(2)},\dots,A^{({N+1})}]\nonumber\\
 &\equiv   \sum^{R}_{r=1} \lambda_r a_{r}^{(1)}\circ a_{r}^{(2)} \circ \cdots \circ a_{r}^{{({N+1})}},\nonumber
 %\label{eq:610}
 \end{flalign}
 and its Tucker decomposition:
 %\begin{small}
 \begin{flalign}
 \mathcal{D}  &\approx [\mathcal{G};A^{(1)},A^{(2)},\dots,A^{({N+1})}]\nonumber\\
 &\equiv  \mathcal{G}\times_1 A^{(1)} \times_2 A^{(2)}\times_3 \cdots \times_{N+1} A^{({N+1})}\hspace{4.2em}\hspace{2.5em}\nonumber\\
 &=\sum^{R_1}_{r_1=1} \sum^{R_2}_{r_2=1}\cdots \sum^{R_{N+1}}_{r_{N+1}=1} g_{r_1 r_2 \dots r_{N+1}} a_{r_1}^{(1)} \circ  a_{r_2}^{(2)} \circ \cdots \circ a_{r_{N+1}}^{({N+1})}.\nonumber%\label{eq:710}
 \end{flalign}
 %\end{small}
 Therefore, analogous to Section \ref{sec:modeling}, we have the
 prediction parameters of the PSR model
 \begin{equation}\label{eq:611}
 %\tilde{m}_{i_1\dots i_{N-1}}(r)=a_{i_1 r}^{(1)} a_{i_2 r}^{(2)} \dots  a_{i_{N-1} r}^{(N-1)},\nonumber\\
 \tilde{m}_{i_1\dots i_{N}}=(\lambda \ast A_{i_1:}^{(1)} \ast A_{i_2:}^{(2)} \ast \cdots \ast A_{i_{N}:}^{({N})})^T,
 \end{equation}
 %\begin{small}
 \begin{equation}\label{eq:712}
 %	\tilde{m}_{i_1\dots i_{N-1}}(r)= \sum^{R_1}_{r_1=1} \sum^{R_2}_{r_2=1}\dots \sum ^{R_N}_{r_N=1} g_{r_1 r_2 \dots r_N} a_{i_1 r_1}^{(1)}   a_{i_2 r_2}^{(2)}  \dots  a_{i_{N-1} r_{N-1}}^{(N-1)}\nonumber ,\\
 \tilde{m}_{i_1\dots i_{N}}=\mathcal{G}_{(N+1)}(A_{i_{N}:}^{({N})}\otimes A_{i_{N-1}:}^{({N-1})} \otimes \dots \otimes A_{i_1:}^{(1)})^{T},
 \end{equation}
 %\end{small}
 via CP and Tucker decomposition methods, respectively.
 The state vector is given by
 \begin{flalign}\label{eq:612}
 x_k = A_{k:}^{({N+1})},
 \end{flalign}
 %The system's updating parameters $M_{i_1\dots i_{N}}$ can be gotten by solving the optimization problem (Eq.~(\ref{eq:68})) after computing $d_{\textbf{ao}}$ and constructing matrix $D_{\textbf{ao}}$, thus we obtain the optimal solution:
 and the transition matrix is
 \begin{flalign}\label{eq:69}
 \tilde{M}_{\textbf{ao}} =  {(X^T X)}^{-1}(X^T D_{\textbf{ao}} X_{\textbf{ao}}) .
 \end{flalign}
 after constructing $X$ and $X_{\textbf{ao}}$, computing $d_{\textbf{ao}}$ and constructing $D_{\textbf{ao}}$.
 
 Algorithm \ref{algo:admmcptd} summarizes the learning procedures. First, we construct the system dynamics tensor  $\mathcal{D}$ from  agents' interaction data set $\Phi_d$~(line 1). We then apply either CP or Tucker decomposition to the system dynamics tensor $\mathcal{D}$, and obtain the prediction parameters $\tilde{m}_{i_1\dots i_{N}}$ through Eq.~(\ref{eq:611}) or~Eq.~(\ref{eq:712})~(lines 2-8). Subsequently, we compute the state vector $x_k$ by Eq.~(\ref{eq:612})~(lines 9-11). For any $\textbf{a}\in \mathcal{A}, \textbf{o}\in \mathcal{O}$, we construct the matrices $X$ and  $X_{\textbf{ao}}$ in Step 3 of Fig. \ref{Fig.CPTUCKER}~(lines 13-16), compute vector $d_\textbf{ao}$ and construct matrix $D_\textbf{ao}$ with $d_{\textbf{ao}}$ on its diagonal~(lines 17-20), and then compute the transition matrix $\tilde{M}_{\textbf{ao}}$ by Eq.~(\ref{eq:69})~(line 21). Finally, we get all the parameters that are needed in order to learn a multi-agent PSR model.
 \begin{algorithm}[!t]
 	\label{algo:admmcptd}
 	\hrule
 	\vspace{0.5em}
 	{Algorithm 1: Tensor-based algorithm for learning multi-agent PSR}
 	\vspace{0.5em}
 	\hrule
 	\LinesNumbered
 	\vspace{0.5em}
 	\small
 	\SetAlgoLined
 	\KwData{number of agents $N$, agents' interaction data $\Phi_d$, decomposition $Methods$~(CP or Tucker), dimension of state compression $R$ (or $(R_{1},R_{2},\dots,R_{N+1})$).}
 	\KwResult{system state vector $x_k$, model parameters $\tilde{M}_{i_1\dots i_{N}}$ and $\tilde{m}_{i_1\dots i_{N}}$.}
 	Construct system dynamics tensor $\mathcal{D}$ from data $\Phi_d$\;
 	\eIf{$Method=CP$}{
 		$ [A^{(1)},A^{(2)},\dots,A^{({N+1})}]\leftarrow CP(\mathcal{D},R)$\;
 		$\tilde{m}_{i_1\dots i_{N}}\leftarrow (\lambda * A_{i_1:}^{(1)} * A_{i_2:}^{(2)}* \cdots * A_{i_{N}:}^{({N})})^T$\;
 	}{
 		$ [\mathcal{G},A^{(1)},A^{(2)},\dots,A^{({N+1})}]= Tucker(\mathcal{D},(R_{1},R_{2},\dots,R_{N+1}))$\;
 		$\tilde{m}_{i_1\dots i_{N}}\leftarrow \mathcal{G}_{(N+1)}(A_{i_{N}:}^{({N})}\otimes A_{i_{N-1}:}^{({N-1})} \otimes \cdots \otimes A_{i_1:}^{(1)})^{T}$\;
 	}
 	\ForEach{$k\in \{1,\dots,n_{N+1}\}$}{
 		$x_k\leftarrow A_{k:}^{({N+1})}$\;
 	}
 	\ForEach{$\textbf{a}\in \mathcal{A}, \textbf{o}\in \mathcal{O}$}{
 		%		 Construct $X$ and  $X_{\textbf{ao}}$ by Step 3 of Fig. \ref{Fig.CPTUCKER}\;
 		$\mathcal{H}^{\prime}_{\textbf{ao}} \leftarrow \{\textbf{h}|\textbf{h} = \textbf{h}^{\prime} \textbf{ao},\textbf{h} \in \mathcal{H},\textbf{h}^{\prime} \in \mathcal{H}\}$\;
 		$\mathcal{H}^{\prime} \leftarrow \{\textbf{h}|\textbf{h} = \textbf{h}^{\prime}\setminus \textbf{ao},\textbf{h} \in \mathcal{H},\textbf{h}^{\prime} \in \mathcal{H}^{\prime}_{\textbf{ao}}\}$\;
 		$X \leftarrow C_{\mathcal{H}^{\prime}}$ \; $X_{\textbf{ao}}\leftarrow  C_{\mathcal{H}^{\prime}_{\textbf{ao}}}$\;
 		$\textbf{t} \leftarrow \textbf{ao}$\;
 		$\tilde{m}_{\textbf{ao}}\leftarrow \tilde{m}_{\textbf{t}}$\;
 		$d_{\textbf{ao}}\leftarrow X\tilde{m}_{\textbf{ao}}$\;
 		$D_{\textbf{ao}}\leftarrow diag(d_{\textbf{ao}}^1,\ldots, d_{\textbf{ao}}^{|\mathcal{H}'|})$\;
 		$\tilde{M}_{\textbf{ao}}\leftarrow {(X^T X)}^{-1}(X^T D_{\textbf{ao}} X_{\textbf{ao}})$\;
 	}
 	%\caption{Tensor-based algorithm for learning multi-agent PSR}
 	\hrule	
 	\vspace{0.5em}
 \end{algorithm}
 \section{Experimental Study}\label{sec:Experiments}
 We implement the prediction models in the platform of MATLAB, and all the computations are conducted on a Windows PC with a 16-core Intel E5-2640 2.60 GHz CPU and 64 GB memory.
 In order to evaluate the learnt PSR models, a series of action-observation sequences (whose length ranges from 1 to 15) are needed to test and evaluate the predictive performance of the models in various problem domains.
 Therefore, we test our approach on four extended versions of standard benchmarks taken from the literatures,
 i.e., Tag \cite{Pineau2003Point}, Gridworld$*$ \cite{Hamilton2014Efficient}, ColoredGridworld$*$ \cite{Hamilton2014Modelling} and Poc-Man$*$ \cite{Silver2010monte-carlo}.
 Moreover, we simply add one more agent in domains Tag and Gridworld$*$ to construct 3-agent systems for testing purpose.
 %Different problem domains have different size of observation space and different size of state space, which represent different uncertainties and randomness of dynamic systems.
 All of them are large domains and were originally defined in a partially observable Markov decision process~({\em POMDP}) \cite{Kaelbling1998planning}.
 
 For every problem domain $d$, agents are given a random exploration strategy to continuously execute actions in the environment to obtain observations. This is to construct the training sample set $\Phi_d $ and the test sample set $\Psi_d $.
 In the training sample set $\Phi_d$, there were 2000 action-observation sequences each of which is with a maximum length of 10~(because some sequences would terminate early, e.g., reaching the target). There are 3000 action-observation sequences each of which has a maximum length of 15 in the test sample set $\Psi_d$.
 
 We conduct 20 experiments (rounds) to evaluate the average performance of each model in every domain. In every test, the single-round test training set used in model training is randomly selected from the training sequence set $\Phi_d$, which has 500 action-observation sequences ( Poc-Man$*$ uses 400 training sequences, considering that it has a relatively large state space), while the single-round test set used in the test has about 1000 action-observation sequences, which is also randomly selected from $\Psi_d$.
 \subsection{Experimental Settings}
 \subsubsection{Problem Domains}
 %\begin{itemize}
 %	\item {\bf Tag.}	
 
 {
 	{\textcircled{\small{1}}} {\bf Tag}

% 	\begin{wrapfigure}{r}{0.5\linewidth}
% 		\centering
% 		\includegraphics[height=5cm, width=5cm,keepaspectratio]{TAG.eps}
% 		\caption{Tag with a medium number of states and observations}
% 		\label{Fig.tag}
% 	\end{wrapfigure}
 	
 	The domain Tag depicted in Fig. \ref{Fig.tag} is a test-bed proposed for multi-agent research \cite{Pineau2003Point, Rosencrantz2003locating}, in which a chasing \textit{Robot} tracks and tags its \textit{Opponent} in an uncertain environment. There is a set of five executable actions $A=\{\textit{North, East, South, West, Tag}\}$ for \textit{Robot}, while $A=\{\textit{North, East, South, West, Noop}\}$ for \textit{Opponent}.
 	%	Each agent can perform any action in the set of all 5 executable actions $A=\{\textit{North, East, South, West, Tag\,(or Nop)}\}$, where \textit{Tag} action is for \textit{Robot} and \textit{Nop} action is for \textit{Opponent}.
 	%Hence, the joint action space is $\mathcal{A}=A\times A$.
 	Each agent receives a -1 bonus for each move.
 	If they are in the same grid, \textit{Robot} can fully observe \textit{Opponent} and should perform $Tag$ action to win a +10 reward; otherwise,  a negative  bonus -10 is returned. The \textit{Opponent} moves away from \textit{Robot} with a chance of 0.8, otherwise stays still. If \textit{Opponent} is tagged, the game is over and \textit{Opponent} will obtain a -10 bonus. After agents have performed any action, each agent can sensor the surrounding in four directions and receive a noisy observation $o_i$ to check whether there are any walls blocking its movement. Therefore, the space of the agents' observation is $O=\{o_1,o_2,\dots,o_{16}\}$.
 	The state space for \textit{Robot} is $S=\{s_1,s_2,\dots,s_{29}\}$ and \textit{Opponent} is $S=\{s_1,s_2,\dots,s_{29},s_{tagged}\}$, which is a set of all possible locations plus with a special tagged state $s_{tagged}$.
 	%$\mathcal{S}$ is represented as the cross product of states $S$ of the two agents, i.e., $\mathcal{S}= S \times S$, where
 	
 		\begin{figure}[!h]
 			\centering
 			\includegraphics[height=5cm, width=6cm,keepaspectratio]{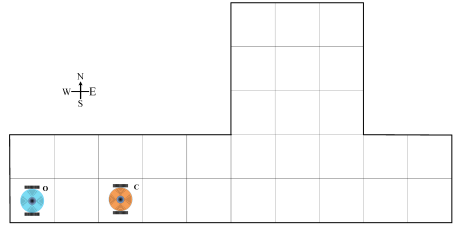}
 			\caption{Tag with a medium number of states and observations.}
 			\label{Fig.tag}
 		\end{figure}
 	
 	{\textcircled{\small{2}}} {\bf Gridworld$*$ and ColoredGridworld$*$}
% 	
% 	\begin{wrapfigure}{r}{0.5\linewidth}
% 		\centering
% 		\subfigure[Gridworld$*$]{
% 			\label{Fig.sub.Gridworld}
% 			%			\includegraphics[height=3.2cm, width=5cm,keepaspectratio]{figure/1/GridWorld.eps}}
% 			\includegraphics[height=3cm, width=5cm,keepaspectratio]{GridWorld.eps}}
% 		\subfigure[ColoredGridworld$*$]{
% 			\label{Fig.sub.GridworldColor}
% 			%			\includegraphics[height=3.2cm, width=5cm,keepaspectratio]{figure/1/GridWorldColor.eps}}
% 			\includegraphics[height=3cm, width=5cm,keepaspectratio]{GridWorldColor.eps}}
% 		\caption{Gridworld$*$ and ColoredGridworld$*$ with a medium number of states and observations. The grid with red circle is the goal}
% 		\label{Fig.GridworldandColor}
% 	\end{wrapfigure}
 	
 	%	\item {\bf Gridworld$*$ and ColoredGridworld$*$.}
 	The domains GridWorld$*$  and ColoredGridWorld$*$ are a direct extension of the domain GridWorld \cite{Hamilton2014Modelling,Hamilton2014Efficient}. All these domains have a 5 $\times$ 12 grid maze (see Fig. \ref{Fig.GridworldandColor}), in which agents must navigate from a fixed start state towards a goal grid.
 	The main difference between them lies in the different responses of the environment to the interactions performed by agents or the different observation that the agents receive.
 	In GridWorld$*$, an agent can sensor the surrounding in four directions and receive a noisy observation to check whether there are any walls blocking its movement, which results in $2^4$ possible observations. While in ColoredGridWorld$*$, the agent can see colored walls with three possible colors. Hence there are 3 possible observations per wall, which results in $2^8$ possible observations in total.
 	In addition, the complexity of colored walls increases the size of observation space exponentially, which results in a huge set of possible tests and histories.
 	The set of all executable actions of each agent is $A=\{\textit{North, East, South, West}\}$.
 	%then the total joint action space is $\mathcal{A}=A\times A$.
 	An agent fails to execute an action with the probability 0.2. If this happens, the agent randomly moves in a direction orthogonal to the specified direction.
 	A reward of 1 is returned at the target state (resetting the environment) and a negative bonus -1 is costed for each move of the agent.
 	The state space of the agent is a set of all possible locations plus with a goal state $s_{52}$, i.e., $S=\{s_1,s_2,\dots,s_{52}\}$.
 	%$\mathcal{S}$ is represented as the cross product of states $S$ of the two agents, where
 		\begin{figure}[!h]
 			\centering
 			\subfigure[Gridworld]{
 				\label{Fig.sub.Gridworld}
 					\includegraphics[height=3.2cm, width=5cm,keepaspectratio]{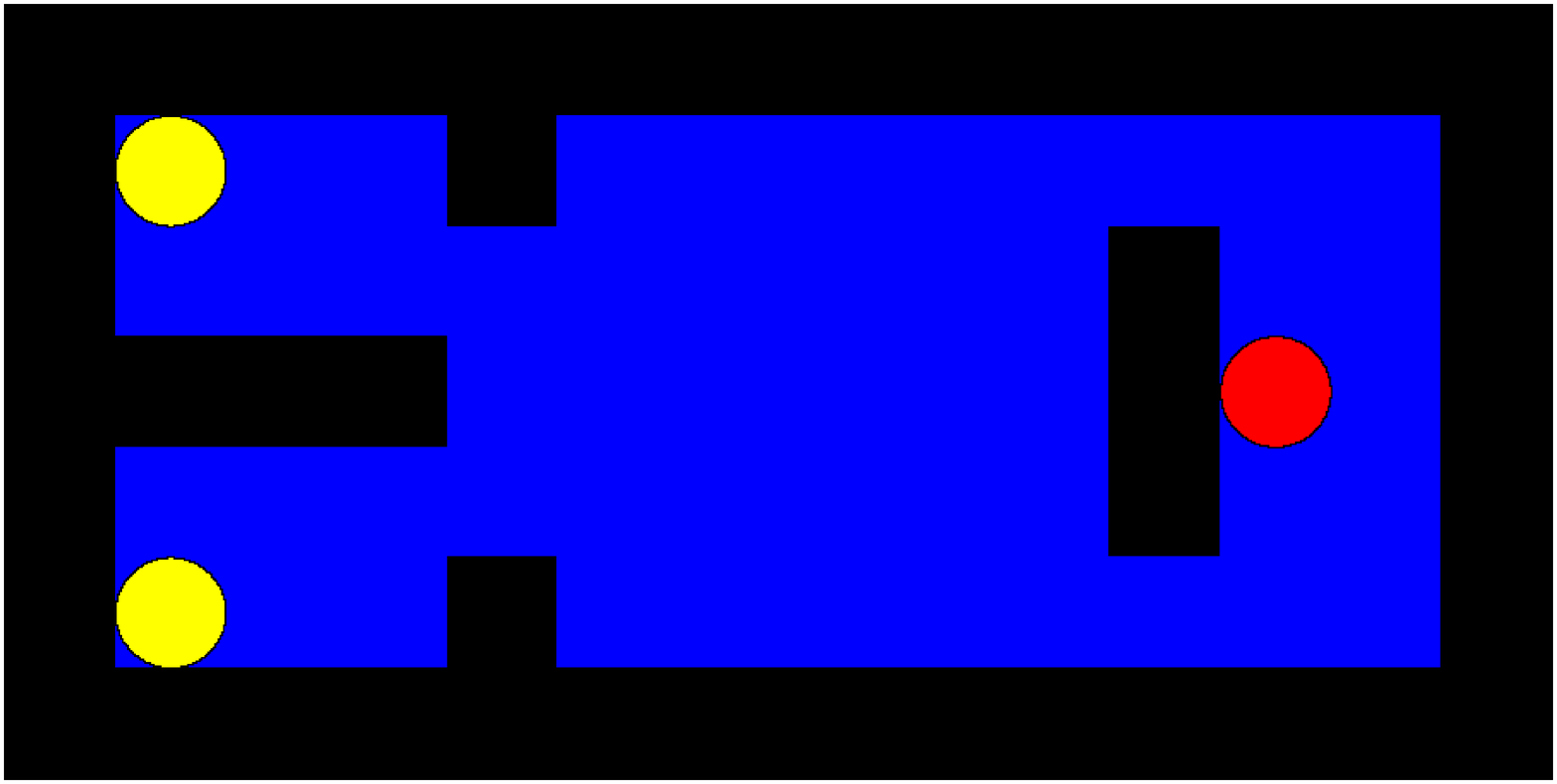}}
 			\subfigure[ColoredGridworld]{
 				\label{Fig.sub.GridworldColor}
 				\includegraphics[height=3.2cm, width=5cm,keepaspectratio]{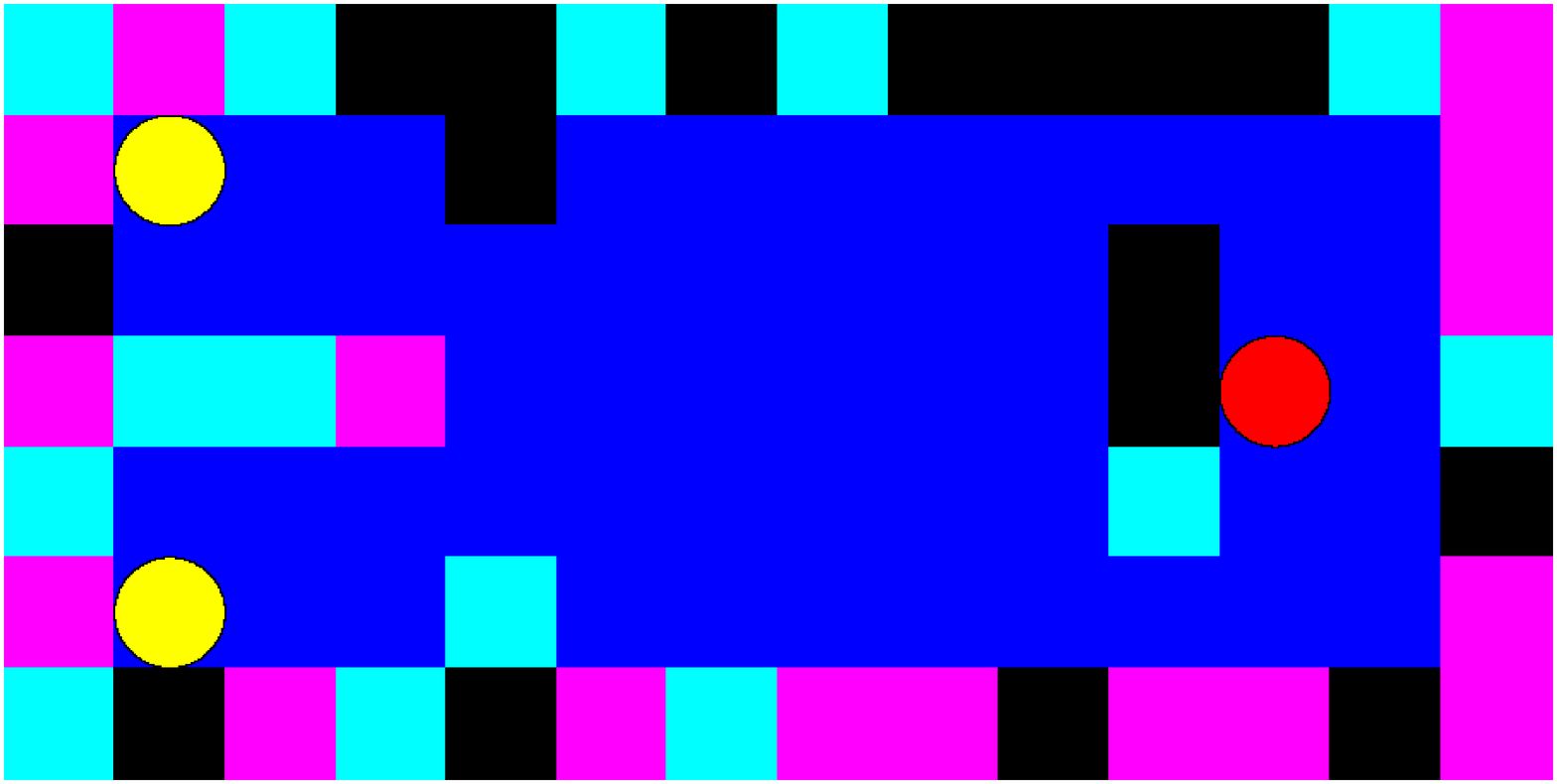}}
 			\caption{Gridworld$*$ and ColoredGridworld$*$ with a medium number of states and observations.The grid with red circle is the goal grid.}
 			\label{Fig.GridworldandColor}
 		\end{figure}
 	%	\item {\bf Poc-Man$*$.}
 	
 	{\textcircled{\small{3}}} {\bf Poc-Man$*$}
 	
% 	\begin{wrapfigure}{r}{0.5\linewidth}
% 		\centering
% 		\includegraphics[height=5cm, width=5cm,keepaspectratio]{pacman_star.eps}
% 		%			\includegraphics[height=6cm, width=6cm,keepaspectratio]{figure/1/pacman_star.eps}
% 		\caption{Poc-Man$*$ with a large number of states and observations}
% 		\label{Fig.pocman}
% 	\end{wrapfigure}
 	
 	The Poc-Man$*$ domain is commonly used for examining the performance of PSR models, which is a variant of the popular video game Pac-Man \cite{Silver2010monte-carlo}. In Poc-Man$*$, the two agents~(marked by yellow points in Fig. \ref{Fig.pocman}) navigate in the maze, gather randomly placed food pellets and keep away from four ghosts~(marked by red points in Fig. \ref{Fig.pocman}) just like in the game scenario Pac-Man. However, in this domain, the agents can only use noisy and partial observations about local environment states to accomplish their mission, which is not identical to the video game version.
 	The set of all executable actions of each agent is $A=\{\textit{North, East, South, West}\}$.
 	%, then the total joint action space is $\mathcal{A}=A\times A$.
 	%And there is a probability of 0.2 when the agent fails to execute a action, which means the action is no deterministic. If agent couldn't success in performing action, the agent randomly moves in a direction orthogonal to the specified direction.
 	An agent fails to execute an action with probability 0.2. If this happens, each agent randomly moves in a direction orthogonal to the specified direction.
 	After the agents have performed any action, they can sensor the surrounding in four directions and receive a noisy observation to check whether there are any walls blocking its movement.
 	
 	%Therefore, the space of agent's observation is $O=\{o_1,o_2,\dots,...\}$ which is a dimension of joint observation space is $\mathcal{O}= O\times O$.
 	Meanwhile, a reward of 1 is returned when each agent finds food pellets and a negative bonus -1 is costed for each move of each agent. Learning a perfect predictive representation of this domain is a challenging task because it has a large size of state space($ \lvert \mathcal{S} \rvert \approx 10^{56}$) and observation space($ \lvert \mathcal{O} \rvert \approx 2^{18}$).
 	
 		\begin{figure}[!h]
 			\centering
 			\includegraphics[height=6cm, width=6cm,keepaspectratio]{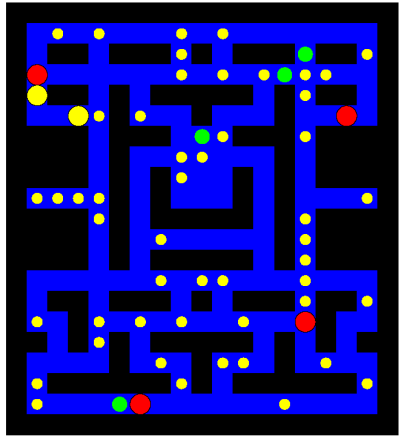}
 			\caption{Pocman$*$ with a large number of states and observations.}
 			\label{Fig.pocman}
 		\end{figure}
 	%\end{itemize}
 	
 	In summary, these domains have different sizes of observation space and state space, which represents different uncertainties and randomness of dynamic systems. In Table \ref{table_SD}, we list the size of the action space, observation space and system state space of each domain for comparative analysis. Meanwhile, we also show the relationship between the agents in each domain.
 	
 	%the joint action space is $\mathcal{A}=A\times A$, and the joint observation space is $\mathcal{O}= O\times O$,
 	%$\mathcal{S}$ is represented as the cross product of states $S$ of the two agents
 	
 	\begin{table*}[t]
 		\centering
 		\caption{Structure of each domain}
 		\vspace{0.5em}
 		\begin{tabular}{|c|c|c|c|c|}
 			\hline
 			% after \\: \hline or \cline{col1-col2} \cline{col3-col4} ...
 			\textbf{Domain}       & \textbf{$|\mathcal{A}|=|A\times A|$} & \textbf{$|\mathcal{O}|= |O\times O|$}     & \textbf{$|\mathcal{S}|= |S\times S|$} & \text{Relationship} \\
 			\hline
 			Tag    & $5\times 5$    & $2^{4}\times2^{4}$ & 870 & Competitive   \\
 			Gridworld$*$   & $4\times 4$    & $2^{4}\times2^{4}$ & 2704 & Competitive   \\
 			ColoredGridworld$*$   & $4\times 4$    & $2^{8}\times2^{8}$ & 2704 & Competitive  \\
 			Poc-Man$*$     & $4 \times 4$   & $ \approx 2^{9}\times 2^{9} $ & $\approx 10^{56}$ & Cooperative   \\
 			\hline
 		\end{tabular}
 		\label{table_SD}
 	\end{table*}
 	
 }
 
 \subsubsection{Comparative Methods}\label{sec:comparativemethods}
 We aim to learn a complete PSR model in different multi-agent systems as elaborated above.
 Note that many algorithms focus on learning a local model of the underlying system with the aim of making only  predictions in specific situations.
 Thus, these algorithms are not included in the comparison.
 For all domains,
 we compare our new learning PSRs techniques~(CP,  NCP, TD and NTD) to traditional methods, i.e.,
 TPSR and compressed PSR (CPSR) approaches \cite{Hamilton2014Modelling,Hamilton2014Efficient}.
 For a fair comparison, we set the compressed dimension of our algorithms~($R$ in CP\,(NCP) and $(P,Q,R)$ in TD\,(NTD)) to be the same as that of TPSR and CPSR algorithms.
 %R=15=P=Q
 
 \subsubsection{Performance Measurements}
 The main purpose of a dynamic system is to predict the probabilities of different observations when executing an action given an arbitrary history.
 We evaluate the learnt models in terms of prediction accuracy,
 which computes
 the gap between the true predictions and the predictions given by the learnt model over all test sequences.
 For each domain, {there are no related POMDP files}, which contains the true value of each step prediction or could obtain through calculating. Hence, we
 cannot obtain the true predictions and use Monte-Carlo roll-out predictions \cite{Hamilton2014Efficient} instead.
 
 The error function  used in our experiments is called absolute error~({\em AE}) that computes the average of absolute error of one-step prediction error per time step given an arbitrary history $\textbf{h}_k$ at time step $s=|\textbf{h}_k|$,
 as shown in Eq.~(\ref{aerror}).
 %And we can easily extend these two kind error function to two step or other steps prediction measurements.
 \begin{equation}
 AE=  \frac{1}{N_T}\sum_{t=1}^{N_T} \lvert  \hat{p}(\textbf{o}_{k+1}^{t}|\textbf{h}_k^{t}\textbf{a}_{k+1}^{t})-p(\textbf{o}_{k+1}^{t}|\textbf{h}_k^{t}\textbf{a}_{k+1}^{t})\rvert.
 \label{aerror}
 \end{equation}
 where $N_T$ is the total number of test sequences~($ N_T=1000 \times 20 $), which is equal to the size of single-round test set times the total number of rounds, and the test of length $k$ starting from 0 to $L-1$~($L=15$)  are used, respectively.
 In Eq.~(\ref{aerror}), $p(\cdot)$ is the probability obtained from the Monte-Carlo roll-out prediction and
 $\hat{p}(\cdot)$ is the estimated probability computed by the learnt model.

 %For each trial of a test data in each environment, using  Eq.~(\ref{rmse}) or Eq.~(\ref{aerror}) we can calculate the predictive margin between learnt model and the underlying system at each time step, then running a statistical analysis we will obtain a full view of the stability and accuracy of our model.
 \subsection{Results}\label{results}

 \begin{figure}
 	\centering
 	\subfigure[Gridworld$*$]{
 		\label{Fig.sub.GridworldEXP04}
 		\includegraphics[height=8cm, width=8cm,keepaspectratio]{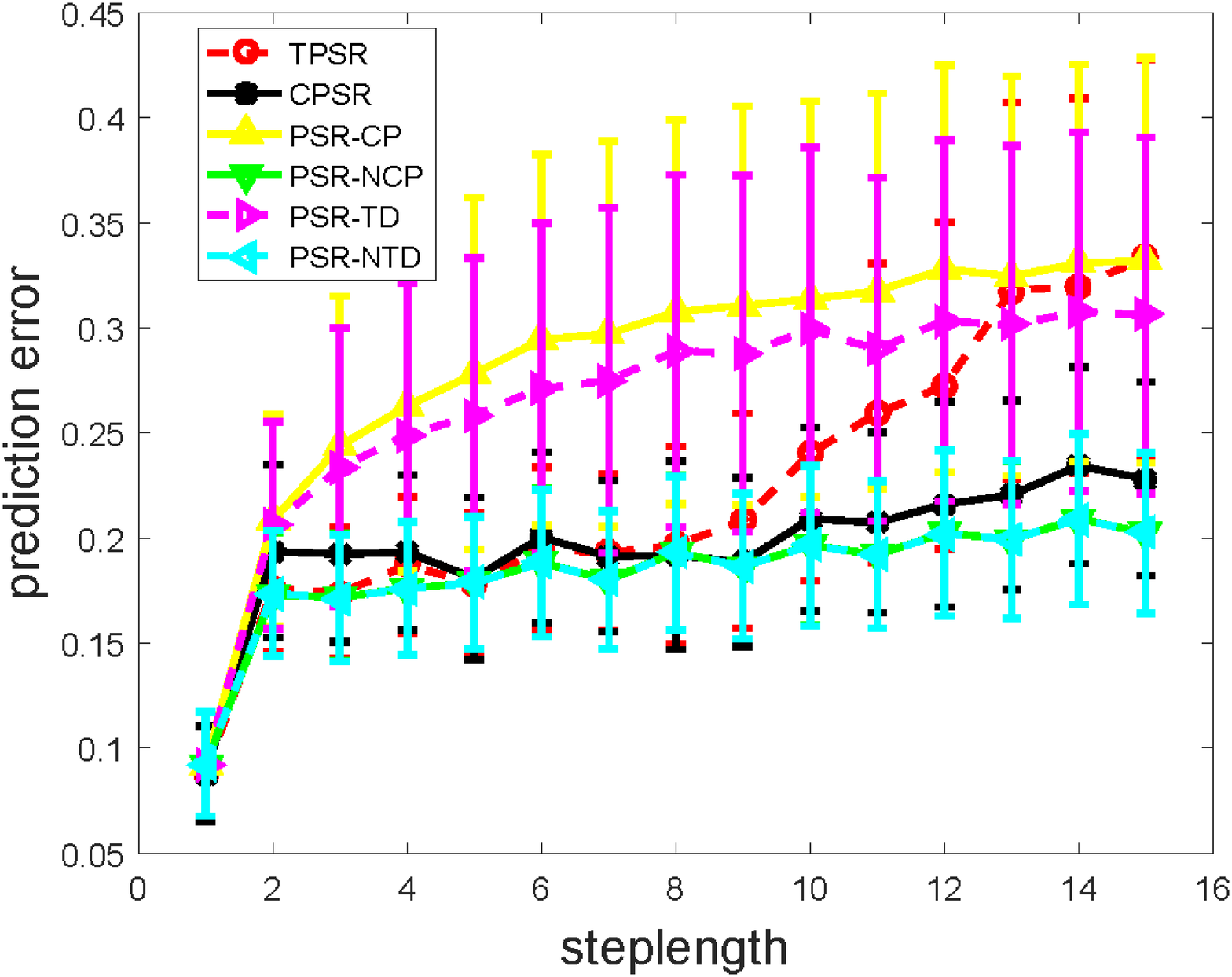}}
 	\subfigure[ColoredGridworld$*$]{
 		\label{Fig.sub.ColoredGridworldEXP04}
 		\includegraphics[height=8cm, width=8cm,keepaspectratio]{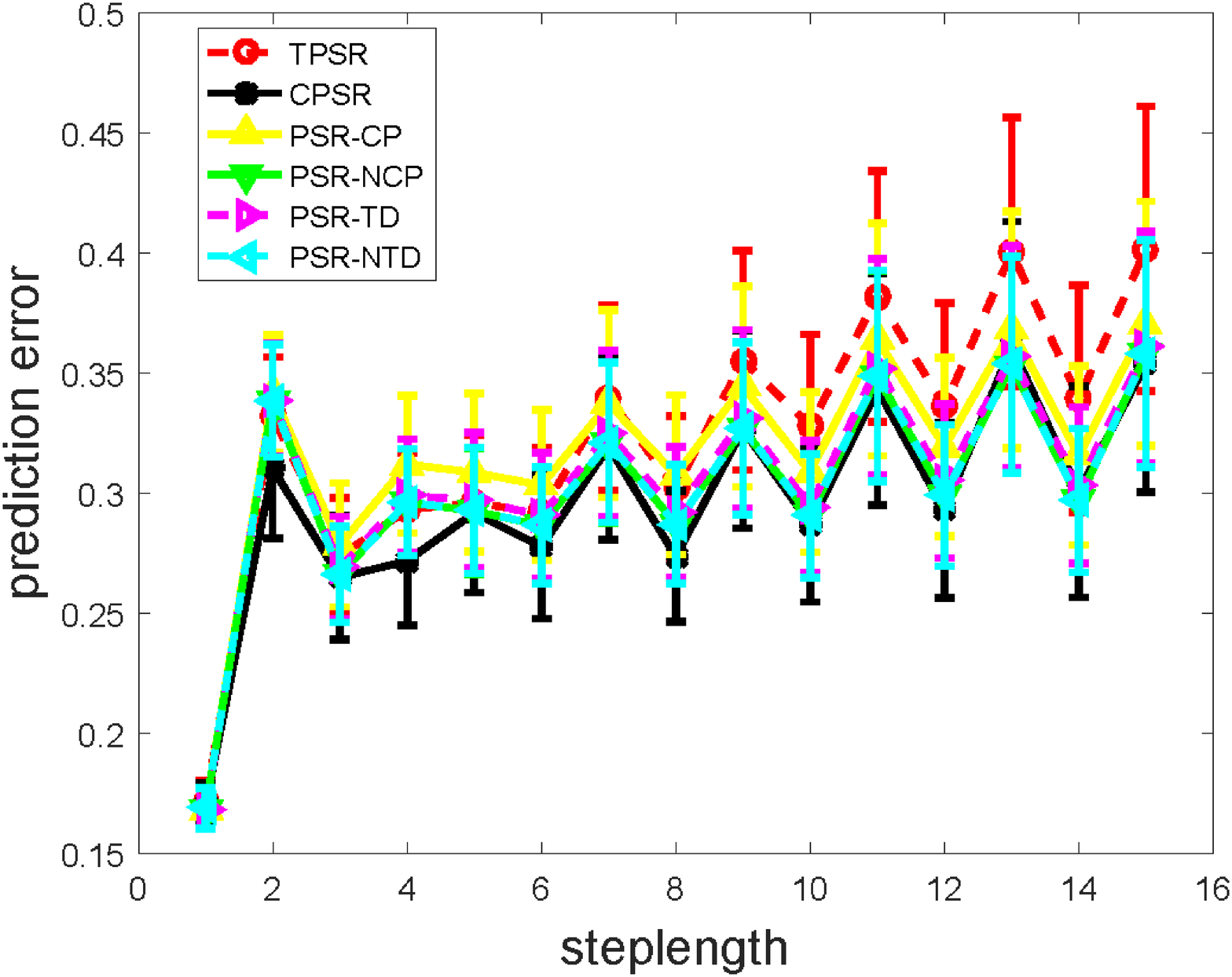}}
 	\subfigure[Poc-man$*$]{
 		\label{Fig.sub.SmallPocmanEXP04}
 		\includegraphics[height=8cm, width=8cm,keepaspectratio]{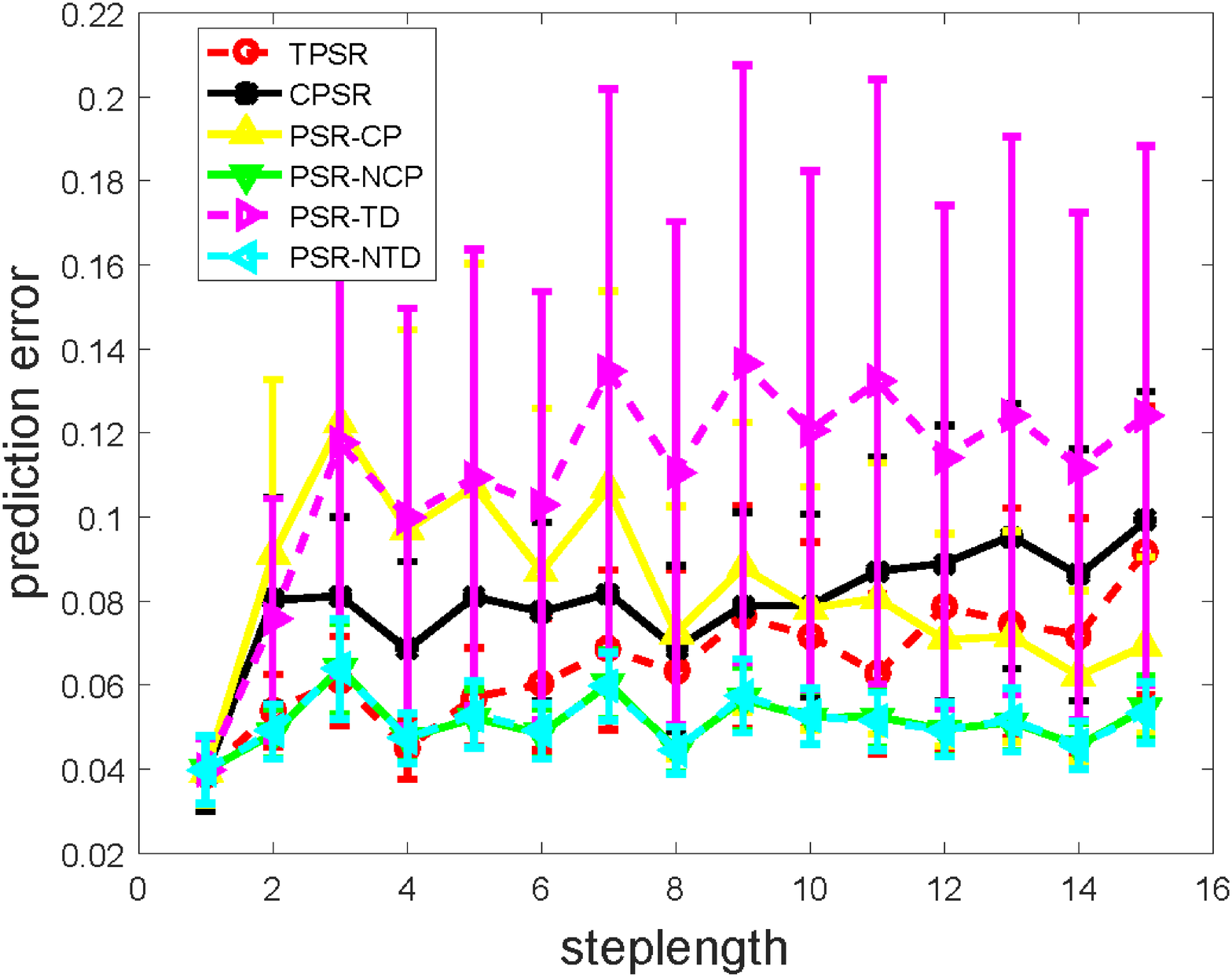}}
 	\subfigure[Tag]{
 		\label{Fig.sub.TagEXP04}
 		\includegraphics[height=8cm, width=8cm,keepaspectratio]{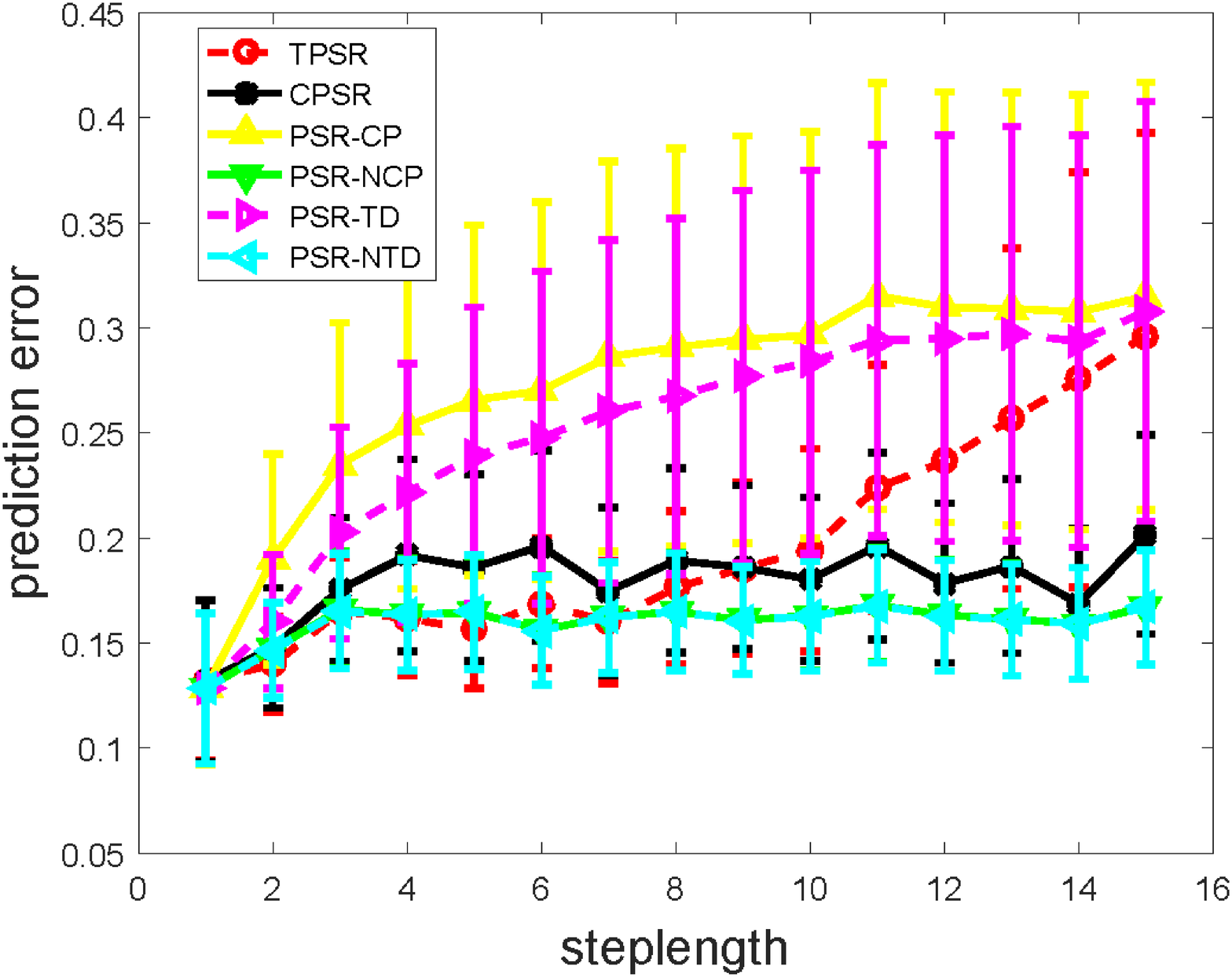}}
 	%	\caption{Comparison analysis of Tensor method, TPSR and CPSR methods in Gridworld$*$, Malmo, Poc-Man$*$ and  Tag}
 	\caption{Comparison analysis of tensor methods, TPSR and CPSR for 2-agent domains Gridworld$*$, GridworldColor$*$, Poc-Man$*$ and Tag}
 	\label{Fig.DomainsMD}
 \end{figure}

 We conduct the experiments to calculate the model accuracy by comparing the one-step prediction accuracy of the evaluated methods in six problem domains~(2-agent domains including Tag, Gridworld$*$, ColoredGridworld$*$, and Poc-Man$*$, and 3-agent domains including Tag and Gridworld$*$), and the average runtime for each domain is also obtained.

 In Fig. \ref{Fig.DomainsMD}, the $x$-axis is the step length of action-observation and the $y$-axis is the mean prediction error of 20,000 trials~($N_T = 1000 \times 20$) calculated by Eq.~(\ref{aerror}).
 As it can be seen from Fig. \ref{Fig.DomainsMD}, for almost all cases in 2-agent system,  our algorithms~(PSR-CP and PSR-TD) perform as well as all the other algorithms, but are not very well when the step-length is bigger than two except ColoredGridworld$*$.
 While PSR-NCP and PSR-NTD perform best and produce more competitive predictions than other algorithms in all horizons for all domains.  As shown in Fig. \ref{Fig.sub.SmallPocmanEXP04}, both PSR-NCP and PSR-NTD algorithm are able to learn more accurate models compared to the  TPSR and CPSR algorithms in the Poc-Man$*$ domain, although the domain is more suitable for the CPSR approach.
 The reason why PSR-NCP and PSR-NTD are technically superior to their competitors is due to the fact that they get a non-negative solution, while CP and TD optimizations do not have a nonnegativity constraint.
 \begin{figure}[t]
 	\centering
 	\includegraphics[height=8cm, width=10cm,keepaspectratio]{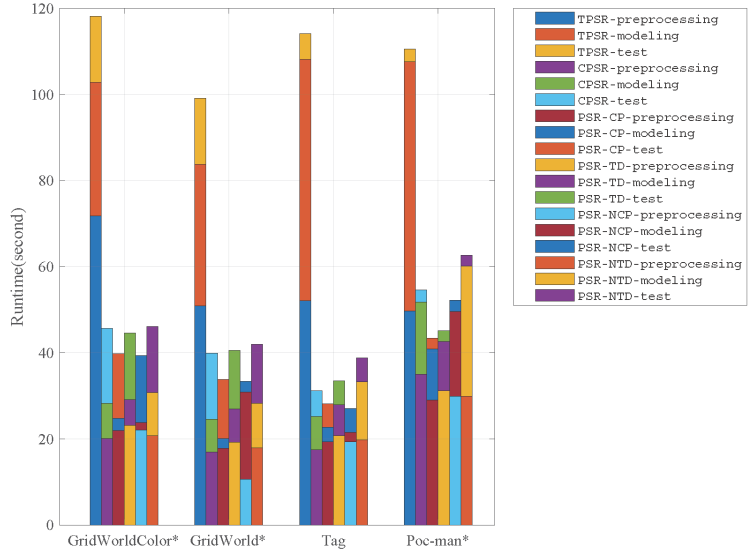}
 	\caption{Runtime in building  and testing the models for four different 2-agent domains}
 	\label{Fig.runtime}
 \end{figure}
 
 The running time of each algorithm is given in Fig. \ref{Fig.runtime}, including the time of three parts:
 
 {\textcircled{\small{1}}} {\em Data preprocessing} includes establishing and normalizing the dynamics matrix~(tensor) of the system and the auxiliary matrices of an algorithm. When an algorithm needs more auxiliary matrices, the computational time will inevitably increase. Especially with the extension of the action-observation sequence or the increasing complexity of a problem domain, the dynamics matrix will eventually become very large. Compared to our methods, TPSR and CPSR methods cost much more time.
 
 {\textcircled{\small{2}}}  {\em Modelling} includes finding the core joint test set of PSR and learning model parameters. TPSR and CPSR need to perform singular value decomposition~(SVD) operations on the dynamics matrix, and our algorithms need to solve tensor decomposition problems. Our methods perform as well as TPSR and CPSR methods in this part. With the benefit of tensor decomposition, we can truly improve the efficiency of the algorithm.
 
 {\textcircled{\small{3}}} {\em Making prediction}. In Poc-Man$*$, our methods spend more time than the others because our methods obtain a  larger set of projection vectors, which  costs much time to the state update of the model when the prediction is carried out.
 
 %\begin{itemize}
 %	\item Data preprocessing: including establishing and normalizing the dynamic matrix~(tensor) of the system and the auxiliary matrix of the algorithm.  When the algorithm needs more auxiliary matrices, the computing time of this part will inevitably increase. Especially with the extension of the action-observation sequence of training data or the increase of the complexity of the problem domain, the dynamic matrix will eventually become very large. Compared with our methods, TPSR and CPSR methods cost much more time.
 %	\item Modelling: including finding the core joint test set of PSR and learning model parameters. TPSR and CPSR need to perform singular value decomposition~(SVD) operations on the dynamic matrix of the system, and our algorithms need to solve tensor decomposition problems. Our method perform as well as TPSR and CPSR methods in this part. With the benefit of tensor decomposition, it can truly improve the efficiency of the algorithm.
 %	\item Making prediction: In Poc-Man$*$, our methods spend much time than the others because our method can obtain a much larger set of projection vectors than other methods, which  costs much time to the state update of the model when the prediction is carried out.
 %\end{itemize}
 
 %}
 \begin{figure}
 	\centering
 	\subfigure[Gridworld$*$~($N=3$)]{
 		\label{Fig.sub.GridworldEXP04m3}
 		\includegraphics[height=8cm, width=8cm,keepaspectratio]{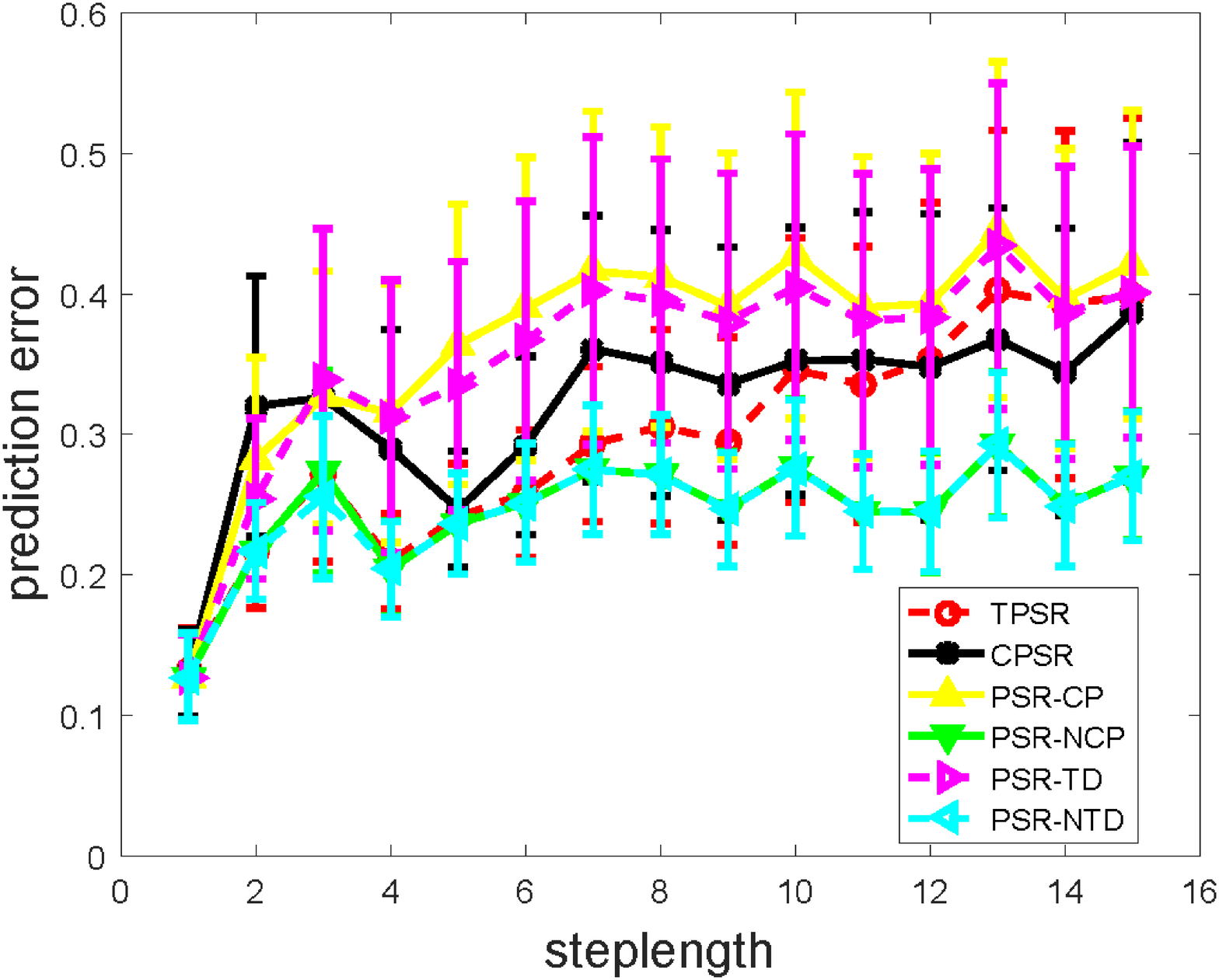}}
 	\subfigure[Tag~($N=3$)]{
 		\label{Fig.sub.TagEXP04m3}
 		\includegraphics[height=8cm, width=8cm,keepaspectratio]{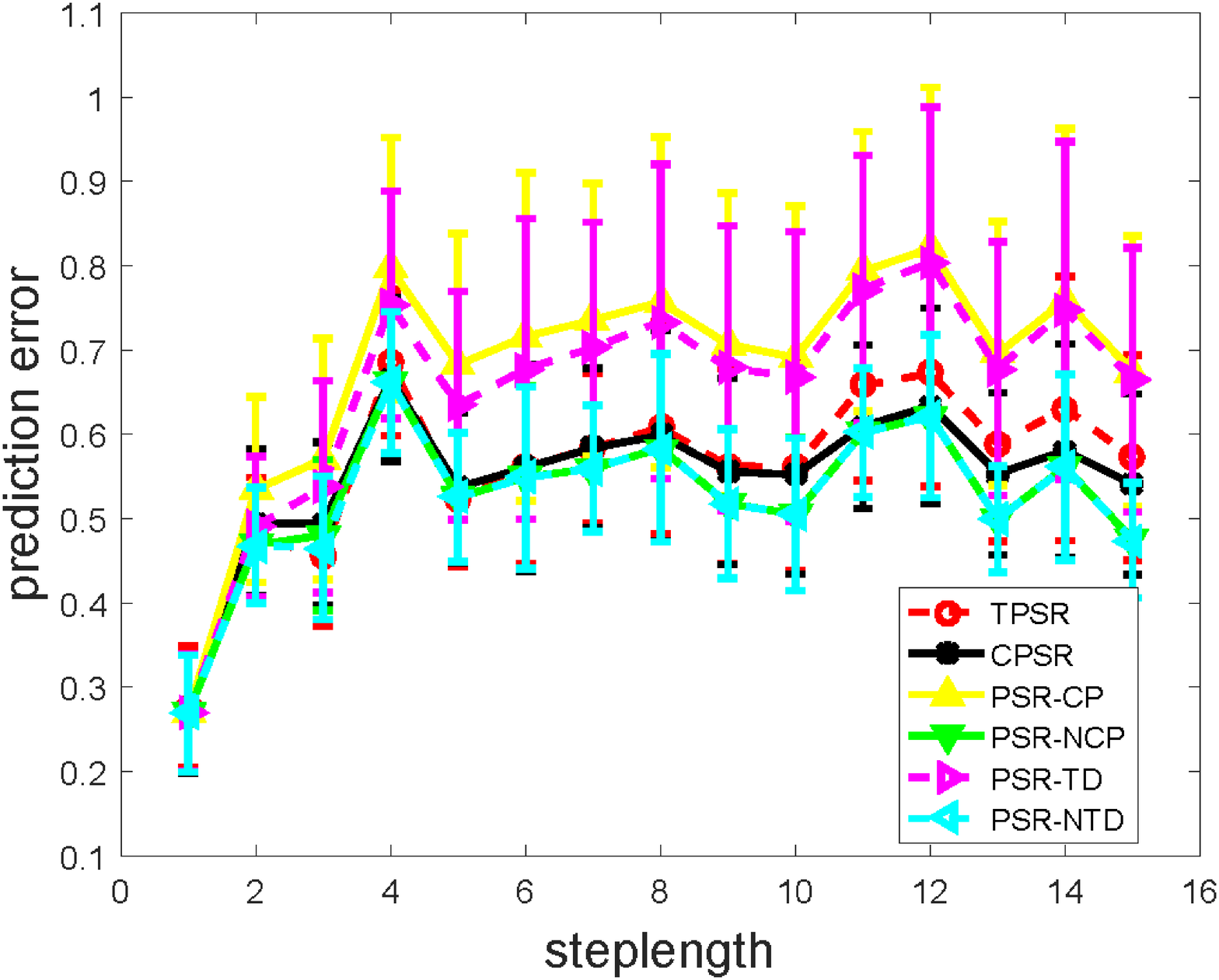}}
 	%	\caption{Comparison analysis of Tensor method, TPSR and CPSR methods in Gridworld$*$, Malmo, Poc-Man$*$ and  Tag}
 	\caption{Comparison analysis of tensor methods, TPSR and CPSR for 3-agent domains Gridworld$*$ and  Tag}
 	\label{Fig.DomainsMDm3}
 \end{figure}
 
 In Fig. \ref{Fig.DomainsMDm3}, we add one more agent for domain Gridworld$*$ and a \textit{Robot} agent for domain Tag. For all horizons of these two domains, PSR-NCP and PSR-NTD perform better than all the other algorithms and produce more competitive predictions. This also show the scalability of our approaches in this article.
 
 In summary, the good performance of our approach is partially due to the fact that the tensor decomposition can dig out the embedded connections of high dimensional data and it is not largely effected by noise in the dynamic system.

 \section{Related Works}\label{sec:relatedwork}
 Predictive state representation~(PSR) represents state of a dynamical system using a function of a vector of statistics about future actions and observations \cite{Littman2001Predictive}. %\citeauthor{Littman2001Predictive}
 Littman et al. \cite{Littman2001Predictive} introduced the PSR principles, theories and modeling methods, and presented a detailed description of the conversion relationship between the PSR models and others. They demonstrated the advantages of PSR models when the models are compared to other traditional approaches,  e.g. POMDPs. After more than a decade of development, most of the PSR research work is devoted to the following four issues:
 the PSR principles,  core test discovery,  PSR model learning and PSR-based planning,
 where the second and third ones are the main interest in this field.
 
 %\citeauthor{Littman2001Predictive}
 Littman et al. \cite{Littman2001Predictive} proposed an algorithm based on sufficient training data for modeling
 the PSR model of the dynamical system. Specifically, this work is based on the assumption that the
 core tests are known,
 and then uses gradient descent method to learn from
 the training data for getting the PSR model.
 Later on, %\citeauthor{McCracken2006Online}
 McCracken et al. \cite{McCracken2006Online} developed constrained gradient descent method, thus the efficiency and accuracy of the PSR model has been improved greatly.
 %\citeauthor{James2004Learning}
 James et al. \cite{James2004Learning} studied a special class of controlled
 dynamic systems with a reset operation and provided
 the first discovery and learning algorithm for PSRs.
 Moreover, %\citeauthor{James2005Combining}
 James et al. \cite{James2005Combining} proposed a model called memory-PSRs and also use landmarks while learning PSRs.
 It can reduce the size of the model (in comparison to a PSR model). In addition, many dynamical systems have memories that can serve as landmarks that completely determine the current state.
 The detection and recognition of
 landmarks is advantageous because they can serve
 to reset a model that has gotten off-track, which
 happens usually when the model is learned from samples.
 However, there are many irrecoverable dynamical systems that cannot be reset in practice.
 For this reason, some researchers are dedicated to non-resettable dynamical systems.
 %\citeauthor{Wolfe2005Learning}
 Wolfe et al. \cite{Wolfe2005Learning} proposed a suffix-history algorithm and a temporal difference algorithm for non-returnable dynamical systems. %\citeauthor{Wiewiora2005Learning}
 Wiewiora et al. \cite{Wiewiora2005Learning} learned PSR from a single sequence~(i.e., history).
 
 %\citeauthor{Rosencrantz2004Learning}
 Rosencrantz et al. \cite{Rosencrantz2004Learning} proposed the transformed PSR~(TPSR),
 which tried to alleviate the discovery problem and learn the parameters of TPSR efficiently by using matrix singular value decomposition for reducing the dimension of system dynamics matrix, and then using the  optimization technology for acquiring the system PSR model.
 In a recent few years, some variants of TPSR were  inspired by this idea
 such as spectral learning approach \cite{Boots2010Predictive,Boots2011An,Boots2010Closing,Kulesza2015LowRank,Kulesza2015Spectral,Rosencrantz2004Learning}, compressed sensing approach \cite{Hamilton2014Efficient,Hamilton2014Modelling}, etc.
 Unlike the traditional iterative methods mentioned before, which can only be used in a toy problem domain,  matrix dimension reduction methods have a quite well performance in practice.
 
 Among these models, researchers often addressed the discovery problem by specifying a
 large set of tests that contains a sufficient subset for state representation.
 %\citeauthor{Hamilton2014Efficient}
 Hamilton et al. \cite{Hamilton2014Modelling,Hamilton2014Efficient} presented compressed transformed
 PSR algorithms for a relatively large domain with a particularly sparse structure.
 Compared to
 TPSR, CPSR allows for an increase in the efficiency and predictive
 power.
 Furthermore, %\citeauthor{Kulesza2015LowRank}
 Kulesza et al. \cite{Kulesza2015LowRank} also did research on data inadequate sampling situation and
 the corresponding algorithm ensures the accuracy of the PSR models in both theory and practice,
 which significantly reduces prediction errors compared
 to standard spectral learning approaches.
 On the other side, there exist other kind of approaches for learning PSR models. %\citeauthor{Kulesza2015Spectral}
 Kulesza et al. \cite{Kulesza2015Spectral} introduced a TPSR-based model with a weighted loss function to
 overcome the consequence of discarding arbitrarily small singular values of the system dynamics matrix.
 They showed that the algorithm can effectively reduce the prediction error within the error bounds; however, the algorithm requires the training data to be sufficiently sampled.
 
 Some researchers learned PSR models using machine learning methods and optimization approaches. %\citeauthor{Liu2015Predictive}
 Liu et al. \cite{Liu2015Predictive} partitioned the entire state space into several  sub-state space and learnt each separate sub-state space via the landmark technique.
 %\citeauthor{Liu2016Learning}
 Liu et al. \cite{Liu2016Learning} formulated the discovery problem as a sequential decision making problem, which can be solved using Monte-carlo tree search.
 %\citeauthor{Huang2018Basis}
 Zeng et al. \cite{zeng2017group} formulated the discovering of the set of core tests as an optimization problem, and then applied alternating direction method of multipliers to solve the problem, which did not require the specification of the number of core tests.
 Huang et al. \cite{Huang2018Basis} proposed a method for selecting a finite set of columns or rows for spectral learning via adopting a concept of model entropy to measure the accuracy of the learnt model. Hefny et al. \cite{hefny2018recurrent} introduced Recurrent Predictive State Policy (RPSP) networks, a recurrent architecture that brings insights from predictive state representations to reinforcement learning in POMDPs environments. Liu et al. \cite{liu2019online} proposed online learning and planning approach for POMDPs domains along with theoretical advantages of PSRs and no prior knowledge of the underlying  system is required. Zhang et al. \cite{zhang2019learning} proposed an algorithm extracts causal state
 representations from recurrent neural networks~(RNNs) for learning state representations, that are trained to predict subsequent observations given the
 history and generalizes PSRs to non-linear predictive models and allows for a formal comparison between generator and history-based state abstractions. Although they have applied many new technologies in the PSR field, they did not extend the PSR of a single-agent scenario to a multi-agent one.
% \begin{figure}
% 	\centering
% 	\fbox{\rule[-.5cm]{4cm}{4cm} \rule[-.5cm]{4cm}{0cm}}
% 	\caption{Sample figure caption.}
% 	\label{fig:fig1}
% \end{figure}
% 
 \section{Conclusion and Future Work}\label{sec:conclusion}
 
 In this paper, by utilizing the concept of tensor, we formulate
 the PSR discovery and learning problem as a tensor decomposition problem.
 With the benefit of tensor decomposition techniques, we extend a single-agent PSR in a multi-agent setting and update the parameters for PSR models as well.
 Experimental results show that
 our method significantly outperforms other popular methods.  Future work would study efficient techniques for learning multi-agent PSRs, i.e., how to develop a more efficient tool for finding a core joint test set from system dynamics tensor through optimization techniques.
 
% \section*{Acknowledgements}
% \noindent
% This work was supported in part by the National Natural Science Foundation of China (Grants No. 61772442 and 61836005). Yifeng Zeng receives the EPSRC New Investigator Award.
 
\bibliographystyle{unsrt}
% \bibliography{template}

\end{document}